\documentclass[hidelinks,10pt,journal,compsoc]{IEEEtran}

\usepackage{graphicx}
\usepackage{multirow}
\usepackage[caption=false]{subfig}
\usepackage{amsmath}
\usepackage{mathtools}
\usepackage{afterpage}
\usepackage[table,xcdraw]{xcolor}
\usepackage{caption}
\usepackage[normalem]{ulem}
\usepackage{xcolor}
\usepackage{hyperref}
\usepackage{cleveref}

\ifCLASSOPTIONcompsoc
  \usepackage[nocompress]{cite}
\else
  \usepackage{cite}
\fi

\begin{document}
%

\title{SwarmTouch: Guiding a Swarm of Micro-Quadrotors with Impedance Control using a Wearable Tactile Interface}

%
%

\author{Evgeny Tsykunov, Ruslan Agishev, Roman Ibrahimov, Luiza Labazanova, Akerke Tleugazy, and Dzmitry Tsetserukou,~\IEEEmembership{Member,~IEEE}

\IEEEcompsocitemizethanks{\IEEEcompsocthanksitem All authors are with the Intelligent Space Robotics Laboratory, Skolkovo Institute of Science and Technology, Moscow, Russian Federation.\protect\\
E-mail: \{Evgeny.Tsykunov, Ruslan.Agishev, Roman.Ibrahimov, Luiza.Labazanova, Akerke.Tleugazy, D.Tsetserukou\}@skoltech.ru}}%

%
%

\markboth{}{}


\IEEEtitleabstractindextext{
\begin{abstract}
To achieve a smooth and safe guiding of a drone formation by a human operator, we propose a novel interaction strategy for a human-swarm communication which combines impedance control and vibrotactile feedback. The presented approach takes into account the human hand velocity and changes the formation shape and dynamics accordingly using impedance interlinks simulated between quadrotors, which helps to achieve a natural swarm behavior. Several tactile patterns representing static and dynamic parameters of the swarm are proposed. The user feels the state of the swarm at the fingertips and receives valuable information to improve the controllability of the complex formation. A user study revealed the patterns with high recognition rates. A flight experiment demonstrated the possibility to accurately navigate the formation in a cluttered environment using only tactile feedback. Subjects stated that tactile sensation allows guiding the drone formation through obstacles and makes the human-swarm communication more interactive. The proposed technology can potentially have a strong impact on the human-swarm interaction, providing a higher level of awareness during the swarm navigation.
\end{abstract}

}

\maketitle

\IEEEdisplaynontitleabstractindextext

%
\IEEEpeerreviewmaketitle

\IEEEraisesectionheading{\section{Introduction}\label{sec:introduction}}

%
%
%
%

\IEEEPARstart{N}{avigation} of quadcopters with a remote controller is a challenging task for many users. In order to intuitively operate a single drone, hand commands were proposed in \cite{Aura}. For immersive drone control with hand gestures, Rognon et al. \cite{Rognon_2018} developed a soft upper body exoskeleton with goggles for the first-person view.
However, it is known that, in many cases, a group of robots can perform much better than a single robot due to its scalability and robustness \cite{Lindsey}.


For many types of missions, autonomous formation flight is suitable. However, for some specific applications, fully or partially guided groups of robots are the only possible solution. The operation of swarm is a significantly more complicated task as a human has to supervise several agents simultaneously. In order for the human to work with the drone formation side by side, robust and natural interaction techniques have to be developed and implemented. Human-swarm interaction (HSI) combines many research topics, which are well described in \cite{Kolling}, and could vary from communication channels to a level of swarm autonomy. The authors in \cite{Cacace_2016} presented a multimodal interaction strategy between a human and a formation of drones for search and rescue operations. Gestures and speech recognition along with a tablet allowed the user to control the fleet of quadrotors.
In this paper, we focus on the interface (control and feedback) between a human operator (leader) and a swarm of robots, addressing the nascent and dynamic field of HSI.

\begin{figure}[t]
\centering
\includegraphics[width=0.49\textwidth]{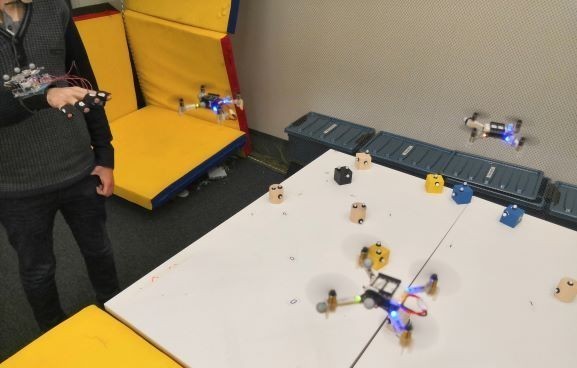}\label{true}
\caption{Human operator manipulates the formation of three quadrotors.}
\label{swarmtouch}
\end{figure}

\begin{figure*}[t]
\centering
\includegraphics[width=\textwidth]{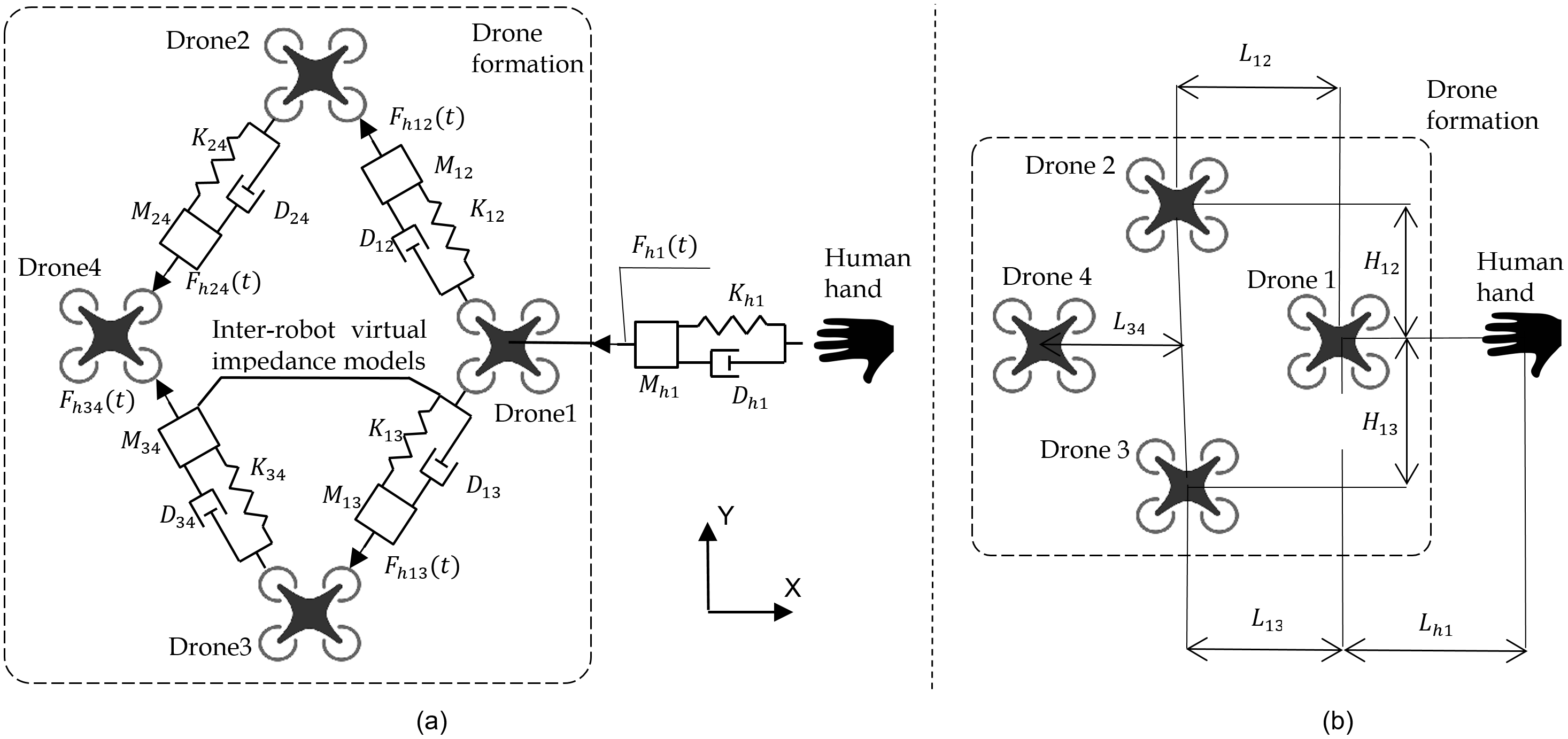}\label{true}
\caption{(a) Position based impedance control, (b) PID position controller. Subscription "h" stands for human.}
\label{imp_architecture}
\end{figure*}

For cases in which a human controls a swarm directly, standard control techniques have been developed in the last few decades. Applications include the interaction between a human and a single-robot or multi-robot systems. Multi-robot formations can be controlled through a central station (centralized control) or, each agent can rely only on local information for making control decisions \cite{Gazi}. To make human-swarm interaction natural and safe, we have developed impedance interlinks between the agents. In contrast to the traditional impedance control \cite{Hogan_1984}, we propose to calculate the external force, applied to the virtual mass of the impedance model, in such a way that it is proportional to the human hand velocity. The impedance model generates the desirable trajectory which reacts to the human hand motion in a compliant manner, avoiding rapid acceleration and deceleration.

The human operator must be aware of changes in the formation (e.g., extension and contraction). The importance of this fact increases with the number of robots. Although visual channels often suffer from poor quality, direct visual feedback or visual information presented with displays \cite{Gioioso_2014} are common ways to deliver information about the formation to the operator.
On the other hand, haptic feedback can also improve the awareness of drone formation state, as reported in \cite{Son_Kim}, \cite{Tsykunov_2018}, and \cite{Stramigioli_2010}. S. Scheggi et al. \cite{Scheggi_2014} proposed a haptic bracelet with vibrotactile feedback to inform an operator about a feasible way to guide a group of ground mobile robots in terms of motion constraints. An arm-worn tactile display for presentation of the collision of a single flying robot with walls was proposed in \cite{Spiss_2017}. Vibrotactile signals improved users' awareness of the presence of obstacles.
Aggravi et al. \cite{Aggravi_2018} developed a wearable haptic display capable of providing a wide range of sensation by skin stretch, pressure, and vibrotactile stimuli. The authors evaluated the proposed device for the control of a fleet of ten simulated quadrotors. Haptic feedback delivered the information about the navigation directions and the connectivity (squeeze) of the fleet. This haptic feedback improved almost all metrics of the experiment.

In contrast to the discussed works, this paper presents a vibrotactile glove for the interaction of the human with a real swarm of aerial robots by providing an intuitive mapping of the formation state to the human finger pads.
It is often easier to estimate the parameters of the whole formation (e.g., dimensions, velocity) rather than to map all environment where the formation is operating. The main novelty of this paper is that we propose to deliver the tactile feedback about the state of the swarm rather than about the distance to obstacles or the desired direction of motion. We designed tactile feedback to convey information about parameters of the formation that are hard to estimate from the visual feedback, i.e., formation state (extension, contraction, and displacement) and state propagation direction (increasing or decreasing drone-to-drone distance). Therefore, tactile cues could effectively supplement the visual channel, making the swarm control more immersive. Cutaneous feedback could play a key role in enhancing the performance of the swarm navigation in the unstructured environment, such as cities.

\section{Formation Control}

\subsection{Approach}

To implement adaptive manipulation of a robotic group by a human operator, such as when the inter-robot distances and formation dynamics change in accordance with the operator state, we propose a position-based impedance control \cite{tsetserukou_tadakuma}.

In our impedance model, we introduce mass-spring-damper links between each pair of agents and between the human and agent formation as shown in Fig. \ref{imp_architecture}(a). The novelty of our impedance model is that we calculate the external force, applied to the virtual mass of each impedance model, in such a way that it is proportional to the operator hand velocity. The target impedance trajectory is processed by PID control which allows high precision positioning and maintains the rhombic shape and orientation of the formation.

While the operator is guiding the formation in space,  impedance models update the goal positions for each flying robot, which changes the default drone-to-drone distances $L_{ij}$, for (i,j=1,2,3,4). As a result, the operator pushes or pulls virtual masses of inter-robot impedance models, which allows the shape and dynamics of the robotic group to be changed by the human hand movement. Each robot relies on the local position information coming from neighbor vehicles and, at the same time, the human operator affects all vehicles globally. Such an adaptive control could lead to a natural multi robot-human interaction.

\subsection{Math} \label{math_section}
In order to calculate the impedance correction term for the robots' goal positions, we solve a second-order differential equation (\ref{imp_eq}) that represents the impedance model. To move in three-dimensional space, we have to solve three differential equations for every impedance link:

\begin{equation}\label{imp_eq} 
    M_d \Delta \ddot{x} + D_d \Delta \dot{x} + K_d \Delta x = F_{ext}(t)
\end{equation}
where $M_d$ is the desired mass of the virtual body, $D_d$ is the desired damping, and $K_d$ is the desired stiffness, $\Delta x$ is the difference between the current $x_{imp}^c$ and desired $x_{imp}^d$ position, and $F_{ext}(t)$ is an external force, applied to the mass.
It is well known that by selecting the desired dynamics parameters for the impedance model, we can achieve various behavior of the oscillator, described by (\ref{imp_eq}), undamped, underdamped, critically damped, and overdamped.
State space representation of (\ref{imp_eq}) has the form:
\begin{equation}
    \begin{bmatrix} \Delta \dot{x} \\ \Delta \ddot{x} \end{bmatrix} = A \begin{bmatrix} \Delta x \\ \Delta \dot{x} \end{bmatrix} + B F_{ext}(t),
\end{equation}
where $A = \begin{bmatrix} 0 & 1 \\ -\frac{K_d}{M_d} & -\frac{D_d}{M_d} \end{bmatrix}$, $B =  \begin{bmatrix} 0 \\ \frac{1}{M_d} \end{bmatrix}$.
In discrete time-space, after integration, we write the impedance equation in the following way:

\begin{equation} \label{eq:imp_integral}
    \begin{bmatrix} \Delta x_{k+1} \\ \Delta \dot{x}_{k+1} \end{bmatrix} = A_d \begin{bmatrix} \Delta x_{k} \\ \Delta \dot{x}_{k} \end{bmatrix} + B_d F_{ext}^k
\end{equation}
where $A_d = e^{AT}$, $B_d = (e^{AT}-I) A^{-1}B$, $T$ is the sampling time, $I$ is the identity matrix, and $e^{AT}$ is the state transition matrix.
The impedance model, as a second order differential equation, can be classified by the shape of the step response, which depends on the poles. The poles are the roots of the characteristic equation:
\begin{equation} \label{characteristic_eq} 
    s^2 + 2 \zeta \omega_n s + \omega_n^2 = 0, 
\end{equation}
\begin{equation} \label{omega} 
    \omega_n = \sqrt{ \frac{K_d}{M_d}}, \zeta = \frac{D_d}{2\sqrt{ M_d K_d }}
\end{equation}
In order to have a critically damped response, $\zeta$ must equal 1. As a result, poles $p_1,p_2$ of (\ref{characteristic_eq}) and the eigenvalues $\lambda_1,\lambda_2$ of matrix $A$ must be equal, real, and positive $\lambda_1=\lambda_2=p_1=p_2$.
The challenging part in (\ref{eq:imp_integral}) is to compute the term $e^{AT}$. The matrix exponential is fined form Cayley-Hamilton theorem, according to which every matrix satisfies its characteristic polynomial. Using those statements, we can find:
\begin{equation}
    A_d = e^{\lambda T} \begin{bmatrix} (1-\lambda T) & T \\ -bT &  (1-\lambda T - aT) \end{bmatrix},
\end{equation}

\begin{equation}
    B_d = -\frac{c}{b} \begin{bmatrix} e^{\lambda T}(1-\lambda T) -1 \\ -bTe^{\lambda T} \end{bmatrix},
\end{equation}
where $\lambda$ is the eigenvalue variable of the matrix $A$, $a = -\frac{D_d}{M_d}$, $b = -\frac{K_d}{M_d}$, $c = \frac{1}{M_d}$. $A_d$ and $B_d$ matrices can be used to calculate the current $x_{imp}^c$ position of the impedance model using equation (\ref{eq:imp_integral}).

To allow the human operator to change the formation shape and dynamics while navigating, the external force term $F_{ext}(t)$ is a function of some human state parameter. We propose to calculate the external force as a function of the human hand's velocity:

\begin{equation} \label{ext_force}
    F_{ext}(t) = K_v v_{h}(t),
\end{equation}
where $K_v$ is a scaling coefficient, which determines the effect of the human hand velocity $v_{h}(t)$ on the formation. To be able to estimate the velocity of the human hand, we make an assumption that it is possible to track the hand motion with some positioning system. During the experimental evaluation we used the Vicon motion capture system.

The method described above is used to calculate the impedance correction vector $\begin{bmatrix}x_{imp}, y_{imp}, z_{imp}\end{bmatrix}^{T}$ or the current position of the virtual body of each impedance model.
In order to demonstrate the performance under assumption on the boundedness of the external inputs, the impedance terms are limited with the maximum values:
\begin{equation} \label{eq:limit_imp}
    \begin{bmatrix} x_{imp} \\ y_{imp} \\ z_{imp} \end{bmatrix} \leq
    \begin{bmatrix}
    x_{imp \textunderscore limit} \\
    y_{imp \textunderscore limit} \\
    z_{imp \textunderscore limit}
    \end{bmatrix},
\end{equation}
where the right side represents the safety thresholds that prevent an overrun of the impedance model.

Finally, the goal positions along $X$, $Y$, and $Z$-axis of each quadrotor are determined as follows (see the structure presented in Fig. \ref{imp_architecture}(a)):

\begin{equation} \label{imp_x}
    \begin{bmatrix} 
    x_{1 \textunderscore g} \\ 
    x_{2 \textunderscore g} \\ 
    x_{3 \textunderscore g} \\ 
    x_{4 \textunderscore g} \end{bmatrix}
    = 
    \begin{bmatrix} x_{h}-L_{h1} \\ x_1-L_{12} \\ x_1-L_{13} \\ \frac{x_2+x_3}{2}-L_{34} \end{bmatrix} 
    -
    \begin{bmatrix} 
    |x_{imp \textunderscore h1}| \\ 
    |x_{imp \textunderscore 12}| \\ 
    |x_{imp \textunderscore 13}| \\ 
    |x_{imp \textunderscore 24}+x_{imp \textunderscore 34}| \end{bmatrix}
\end{equation}

\begin{equation} \label{imp_y}
    \begin{bmatrix} 
    y_{1 \textunderscore g} \\ 
    y_{2 \textunderscore g} \\ 
    y_{3 \textunderscore g} \\ 
    y_{4 \textunderscore g} \end{bmatrix} 
    = 
    \begin{bmatrix} y_{h} \\ y_1 + H_{12} \\ y_1 - H_{13} \\ \frac{y_2+y_3}{2} 
    \end{bmatrix}
    + 
    \begin{bmatrix}
    y_{imp \textunderscore h1} \\
    y_{imp \textunderscore 12} \\
    y_{imp \textunderscore 13} \\
    y_{imp \textunderscore 24}+y_{imp \textunderscore 34} \end{bmatrix}
\end{equation}

\begin{equation} \label{imp_z}
    \begin{bmatrix} 
    z_{1 \textunderscore g} \\ 
    z_{2 \textunderscore g} \\ 
    z_{3 \textunderscore g} \\ 
    z_{4 \textunderscore g} \end{bmatrix} 
    = 
    \begin{bmatrix} z_{h} \\ z_1 \\ z_1 \\ \frac{z_2+z_3}{2}
    \end{bmatrix}
    + 
    \begin{bmatrix}
    z_{imp \textunderscore h1} \\
    z_{imp \textunderscore 12} \\
    z_{imp \textunderscore 13} \\
    z_{imp \textunderscore 24}+z_{imp \textunderscore 34} \end{bmatrix}
\end{equation}
where $x_{imp \textunderscore ij}$, $y_{imp \textunderscore ij}$, and $z_{imp \textunderscore ij}$ for $i,j=h,1,2,3,4$ are corresponding impedance correction terms, $L_{ij}$ for $i,j=h,1,2,3,4$ are displacements for the quadrotors, as could be seen in Fig. \ref{imp_architecture}(b), and $x_i, y_i, z_i$ for $i=1,2,3,4$ are the actual positions of UAVs.

\Crefrange{imp_x}{imp_z} consist of two parts. The first part determines the default geometrical shape of the formation (rhombus which is placed in the $XY$ plane in this case), and the second describes the impedance interlinks between the agents. Both parts of the equations are independent and could be designed separately, following the specific application needs. Although in this paper we consider the rhombic shape, the formation could have an arbitrary geometry, which is defined in the first part of (\ref{imp_x}) to (\ref{imp_z}). The number of UAVs also could be arbitrary. Given some shape, the impedance connections could be designed in such a way, that they do not have to replicate the geometry. We select the impedance links based on the behavior we want to achieve. For example, if we want the distance between Drone 2 and Drone 3 to increase when the formation is moving in the $Y$ direction, then we could introduce an additional impedance interlink between Drone 2 and 3, see Fig. \ref{imp_architecture}.

One of the ideas behind the proposed impedance control is safer operation when the distances between the agents are increasing with increasing velocity. In particular, when the formation is moving fast, we want the drones to always split apart in the negative direction of the $X$ axis (from the human), that is why we subtract the absolute values of impedance terms in (\ref{imp_x}). On the other hand, considering motion in the $Y$ and $Z$ axes, the formation has to be shifted in different directions, with respect to the human motion. If the human starts to move in the left direction, the robotic swarm, following the human, has to shift to the right, demonstrating a "tail" behavior, as shown in Fig. \ref{tail}. Based on the presented discussion, we conclude that the developed impedance control introduces a directional behavior - from the human to the last quadrotor. Thus, the set of impedance links represents a connected directional graph.

As can be seen in (\ref{imp_x}) to (\ref{imp_z}), each agent relies on local information about the distances to neighbor vehicles (geometrical part of the equations), and at the same time on the state of local impedance models. The human affects all impedance interlinks globally.
The computation could be done onboard or on the ground station with corresponding advantages and disadvantages of both approaches.
For the experiment, we used the ground-based Linux Computer to compute all trajectories for each drone in real time. Each drone received its next waypoint through the radio link (communication bandwidth increases linearly with the number of robots).
Decentralized onboard computation is also an option, which requires additional setup.
The velocity of the human hand has to be measured and broadcasted to each drone using a simplex radio channel. We used a Vicon motion capture system for human positioning. Alternative approaches to tracking the position of the human hand which allow for more varied applications are discussed in Section 6.
In addition to the position of the human hand, each agent has to know the distances to the neighbors according to (\ref{imp_x}-\ref{imp_y}). The experimental setup in this paper does not support measuring relative distances with onboard sensors. However, it is possible to achieve this with vision-based methods as shown in \cite{Petracek}. Another option is to set up communication channels between the closest neighbors.
Computational complexity coming from (\ref{imp_eq}) increases linearly with the number of impedance links.

\subsection{Verification} \label{verification_section}
We used a formation of four Crazyflie 2.0 quadrotors to perform the verification flight tests.
To get the high-quality tracking of the quadrotors and human glove during the experiments, we used Vicon motion capture system with 12 cameras (Vantage V5) covering a 5 m × 5 m × 5 m space. We used the Robot Operating System (ROS) Kinetic framework to run the development software and ROS stack \cite{Honig_2015} for Crazyflie 2.0. The position and attitude update rate was 60 Hz for all drones. Before conducting any type of experiment, we ensured that we were able to perform a stable and smooth flight, following the desired trajectory. In order to do so, all PID coefficients for position controller were set to default values for Crazyflie 2.0, according to \cite{Honig_2015} (for x,y-axis $k_p$=40, $k_d$=20, $k_i$=2; for z-axis $k_p$=5000, $k_d$=6000, $k_i$=3500).

As a preliminary experiment, the selection of the impedance parameters was carried out.
First, $M_d$, $D_d$, and $K_d$ coefficients of the impedance model were set in order to get a critically damped response, which would be smooth and comfortable for a human.
To archive this, $\zeta$ must equal 1 in (\ref{characteristic_eq}). Therefore, based on (\ref{omega}), the following condition has to be satisfied $D_d^2 - 4 K_d M_d = 0$.
Making sure that it is true, we selected arbitrary desired dynamic coefficients ($M_d=1.9, D_d=12.6, K_d=21.0$).

Second, human velocity coefficient $K_v$, used for force calculation in (\ref{ext_force}), was selected. We assume that the impedance correction of the goal position has to be no more than 30-50\% of the distances to the neighbors $L_ij$ and $H_ij$ (which is 0.5 meters in this case). We also estimated that the normal human hand velocity, which was estimated from a set of consecutively measured positions provided by a motion capture system, does not go over $1.5 m/sec$ while manipulating the formation. Based on this,  we selected $K_v$ to be $-7 N sec/m$. A negative $K_v$ value is used because when the human is moving in one direction, drones retreat towards the opposite direction (see Section \ref{math_section}).
Finally, for safety reasons, we set the threshold limit of impedance correction term $x_{imp \textunderscore limit}$ to be 0.25 meters for the experiments. For simplification purposes, we used the same dynamic parameters for all impedance models in the experiment.

\begin{figure}[t]
\centering
\includegraphics[width=0.49\textwidth]{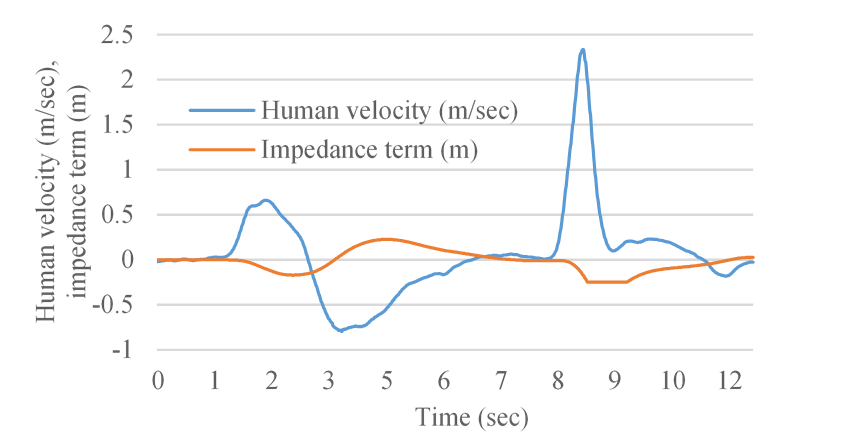}\label{true}
\caption{Human hand velocity (blue) and impedance correction term (orange) versus time. Movement is along $Y$-axis.}
\label{vel_vs_impedance}
\end{figure}

After the selection of all impedance parameters, we checked the single drone behavior, while being guided by the human operator with the proposed impedance controller. To do so, we took Drone 1 and the human wearing a glove, as seen in Fig. \ref{imp_architecture}. We present the values along $Y$-axis. Human hand velocity $v_{h}(t)$ used in (\ref{ext_force}) and the impedance correction term $y_{imp \textunderscore h1}$ used in \ref{imp_y} are shown in Fig. \ref{vel_vs_impedance}. From Fig. \ref{vel_vs_impedance} it can be seen that the impedance model changes its state smoothly in accordance with human hand movement. Due to the negative velocity coefficient $K_v$, human velocity, and impedance term are moving in opposite directions. It is also possible to notice (for the time range 8.5-9 seconds in Fig. \ref{vel_vs_impedance}), that the safety threshold $y_{imp \textunderscore limit}$ helps to prevent dangerous behavior due to high values of the input parameter (human velocity $v_{h}(t)$).

\begin{figure}[t]
\centering
\includegraphics[width=0.49\textwidth]{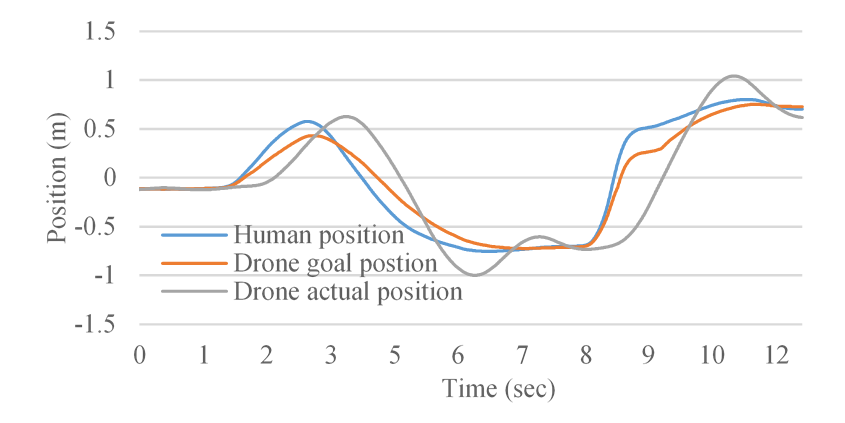}\label{true}
\caption{Human hand position while guiding the drone (blue), drone goal position while following the human (orange), drone actual position (gray) versus time. Along $Y$-axis.}
\label{h_vs_drone1}
\end{figure}

Fig. \ref{h_vs_drone1} shows the actual position of the human hand along with goal and actual positions of Drone 1 (along $Y$-axis). According to Fig. \ref{imp_architecture}(b), $Y$-coordinates of the human and Drone 1 goal position have to be equal, in the case of a simple PID controller. However, due to the impedance correction of the goal position in (\ref{imp_y}), in Fig. \ref{h_vs_drone1} it can be seen that the Drone 1 goal position is slightly behind the human position (this difference is equal to the impedance term $y_{imp \textunderscore h1}$). The result could be represented as a sort of filtering of the robot goal position, which leads to smoother drone guidance, especially in the case of extreme external inputs. 
Afterward, the Drone 1 goal position is provided to the positional PID controller. 
A delay occurs between a human command and a drone reaction, which is expected due to the nature of the impedance controller.

\begin{figure}[t]
\centering
\includegraphics[width=0.49\textwidth]{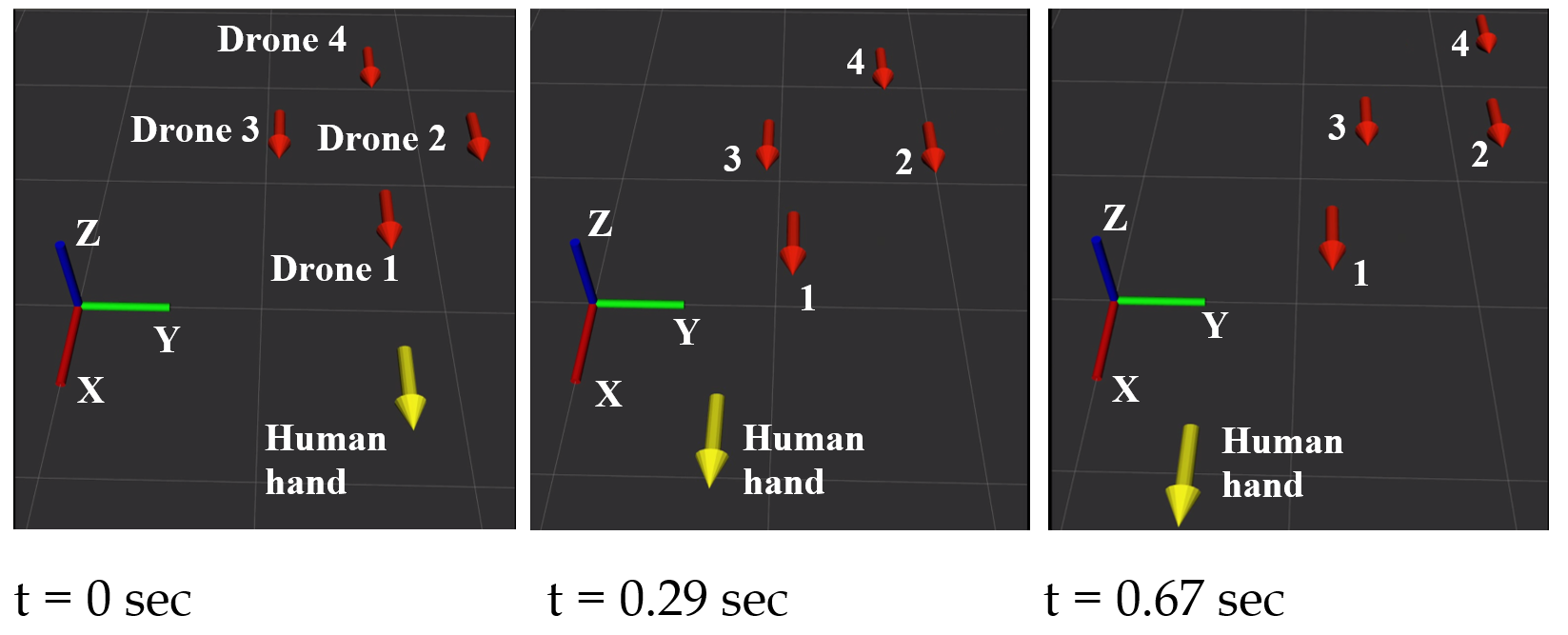}\label{true}
\caption{Formation of four drones (red arrows) following a human hand (yellow arrow). The beginning of the yellow arrow represents the human’s actual position and the beginning of red arrows represents the quadrotor goal positions. The orientation of the arrows represents the orientation of drones and the human hand. The magnitude of the arrows has no significance.}
\label{tail}
\end{figure}

\begin{figure}[t]
\centering
\includegraphics[width=0.49\textwidth]{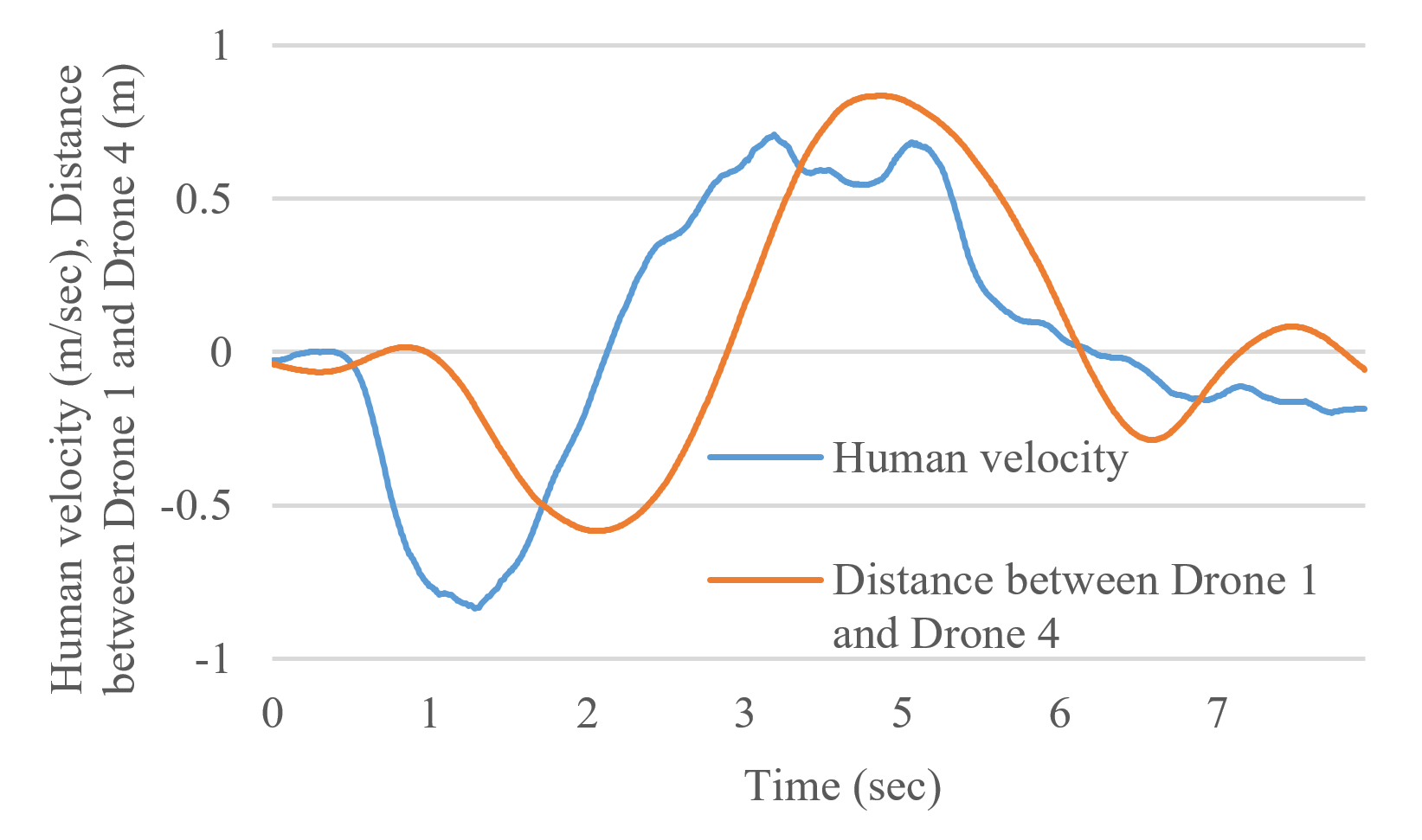}\label{true}
\caption{The blue line represents human hand velocity versus time, the orange line represents distance between Drone 1 and Drone 4. Along $Y$-axis.}
\label{human_vs_distance_1_4}
\end{figure}

\begin{figure}[t]
\centering
\includegraphics[width=0.49\textwidth]{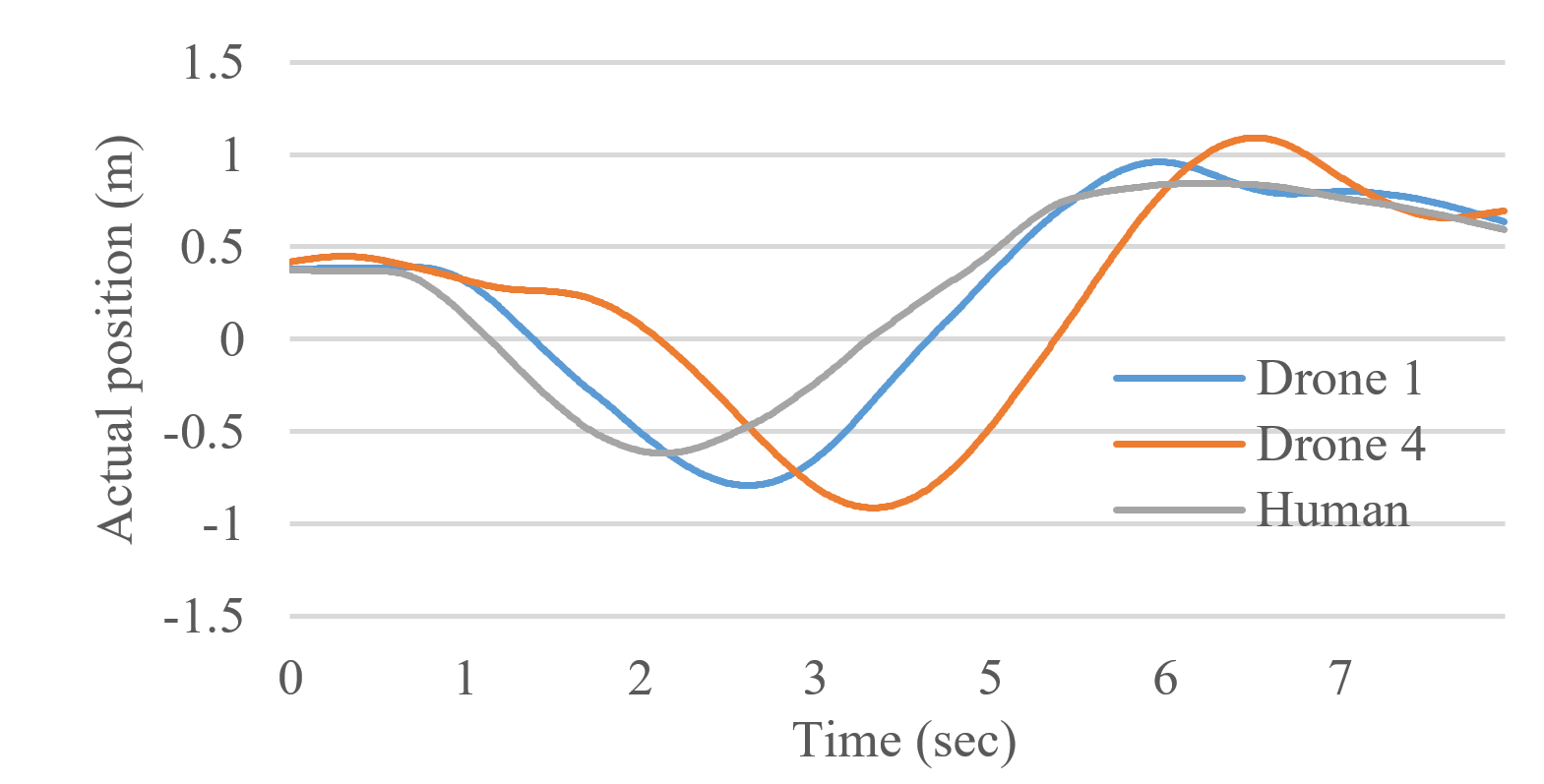}\label{true}
\caption{Actual positions of Drone 1 (blue), Drone 4 (orange), human hand (gray) versus time. Along $Y$-axis.}
\label{human_vs_drone_1_4}
\end{figure}

The next step is to demonstrate the performance of the proposed algorithm for the formation guidance.
We refer to Fig. \ref{tail}, where a human guides four quadrotors with the control structure presented in Fig. \ref{imp_architecture}(a). An operator moves their hand towards the negative $Y$ and the positive $X$ direction. This figure presents an interesting feature of the impedance control, which was mentioned before in Section 2.2. While the human starts to move fast enough, the formation immediately spreads along the direction, which is opposite to the human motion. When the human hand velocity starts to decrease, the formation contracts back to its initial shape. The axis, along which the formation changes its shape, coincides with the human velocity vector.
Fig. \ref{human_vs_distance_1_4} shows the distance along the $Y$-axis between Drone 1 and Drone 4, which are placed in accordance with Fig. \ref{imp_architecture}(a). The displacement between drones is presented in Fig. \ref{human_vs_drone_1_4}. Here could be seen that the magnitude of the displacement is increasing for drones farther away from the human. To address this issue, in future work, the stability of the proposed method will be considered with the increased number of drones.
The proposed control algorithm could be used not only for human-swarm interaction (HSI) but also for obstacle/collision avoidance.

\subsection{Collision Avoidance}
Apart from internal factors that affect the swarm state, such as mass-spring-damper links between the drones, there also could be external reasons which could cause the formation to change, e.g., obstacles. We assume that, within the swarm, every agent decides where to go next using both the local information about surroundings and the global goal (direction and velocity of motion). In this scenario, each quadrotor can plan its obstacle avoidance while considering the position of the nearest obstacles and neighbor agents. The planning algorithm is described below.

The location of drones and obstacles is defined by a Vicon motion capture system, as described in Section \ref{verification_section}. Each quadrotor is aware of the position of local obstacles. Additionally, each obstacle has a safety zone around its center, which is defined as a cylinder (a circle for planar motion) with the predefined radius.

Every controlled robot in the swarm should not only be aware of static obstacles on the map but also take into account moving obstacles, such as humans and other agents in the formation. Collision avoidance method based on the artificial potential field method, \cite{Khatib86}, was applied in this paper to ensure safe real-time robots swarm navigation in a dynamic environment.
The basic idea of the obstacle avoidance algorithm is to construct a smooth function over the extent of robot's configuration space which has high values when the robot is near to an obstacle and lower values when it is further away. This function should have the lowest value at the desired location of the robot. If such a function is constructed, its gradient can be used to guide the drone to the goal configuration. Typically this function consists of two components, attractive and repelling.

In our case, the artificial potential affects a robot's motion in $X$- and $Y$- directions.
An attractive potential function, $U_a(x,y)$, can be constructed by considering the distance between the current position of the robot, $\boldsymbol{p} = [x,y]^T$, and the desired goal location, $\boldsymbol{p}_g = [x_g, y_g]^T$, as follows:
\begin{equation}\label{attractive_potential}
    U_a(x,y) = \xi || \boldsymbol{p} - \boldsymbol{p}_g ||^2
\end{equation}
Here $\xi$ is the constant scaling parameter.


A repulsive potential function in the plane, $U_r(x,y)$, can be constructed based on the distance, $\rho(x,y)$, to the closest obstacle from a given point, $[x,y]$, in configuration space.
\begin{equation}
    U_r(x,y) =
    \begin{cases}
      \eta (\frac{1}{\rho(x,y)} - \frac{1}{d0})^2 & \quad \textrm{if} \quad  \rho(x,y) < d0 \\
      0 & \quad \textrm{if} \quad  \rho(x,y) \geq d0
    \end{cases}
\end{equation}
Here $\eta$ is simply the constant scaling parameter, and $d0$ is a parameter that defines the influence radius of the repulsive potential.

Once the combined potential, $U(x,y) = U_a(x,y) + U_r(x,y)$ is constructed,
a robot's desired velocity can be estimated as $\boldsymbol{v} \propto -\nabla U(x,y)$.

In our case of the human-guided swarm, a point of attraction, $\boldsymbol{p}^d_g$, (goal location)  is assigned to every drone, $d$, relative to the leader-drone position with a prescribed geometrical shape.





HSI could significantly benefit if we couple the described control methods with tactile feedback, forming an interface (control and feedback) between a human and a formation. Informing a human operator about the dynamic formation state (extension or contraction, for example) at the current time could potentially improve controllability.

\section{SwarmGlove: Vibrotactile Wearable Glove}

\subsection{Technology}
The wearable tactile displays, e.g., LinkTouch can represent multimodal information at the fingertips, i.e. force vector, vibration, and contact state \cite{Tsetserukou_LinkTouch}. However, vibration motors, which are easy to control, are widely applied in Virtual Reality \cite{Martinez_2016}, \cite{Maereg_2017}. We applied eccentric rotating mass (ERM) vibration motors which deliver the dynamic state of the swarm in the form of tactile patterns.

\begin{figure}[t]
\centering
\includegraphics[width=0.49\textwidth]{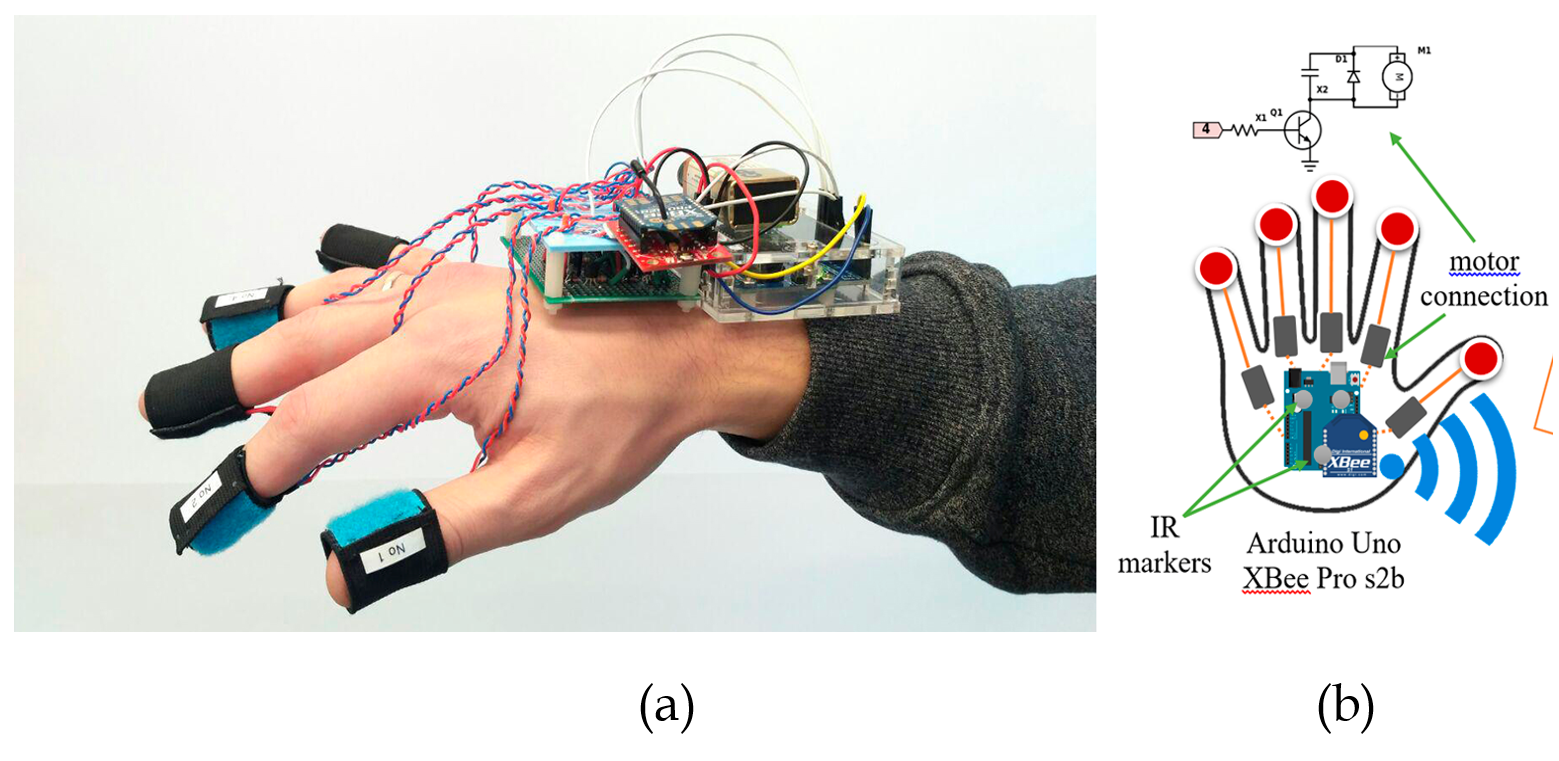}\label{true}
\caption{(a) - wearable tactile display, (b) - tactile device diagram.}
\label{glove}
\end{figure}

We have designed a prototype of the tactile display with five ERM vibrotactile actuators attached to the fingertips, as shown in Fig. \ref{glove}(a). The vibration motors receive control signals from an Arduino UNO controller. The unit with Arduino UNO and battery are worn on the a wrist as a portable device. Infrared reflective markers are located on the top of the unit. The frequency of vibration motors is changed according to the applied voltage. The haptic device diagram is shown in Fig.\ref{glove}(b). The glove microcontroller receives values of the formation state parameters from the PC. The Bluetooth and USB communications between the computer and haptic device were presented in the previous research of the authors in \cite{Tsetserukou_LinkTouch}. The approach in \cite{Tsetserukou_LinkTouch} is limited in working distance and mobility. Therefore, we implemented a radio frequency connection through XBee Pro s2b radio modules due to its robustness and high speed of data exchange. After the Arduino UNO gets the information about the current swarm state, it applies an appropriate vibration pattern.

\subsection{Tactile Patterns}
We designed eight tactile patterns for presenting the feeling of the swarm behavior at the operator's fingertips. Our motivation for the selection of the particular tactile pattern was to bring valuable information that potentially can improve the quality (speed, safety, precision) of operation of the swarm in a complex outdoor environment.

\begin{figure}[t]
\centering
\includegraphics[width=0.49\textwidth]{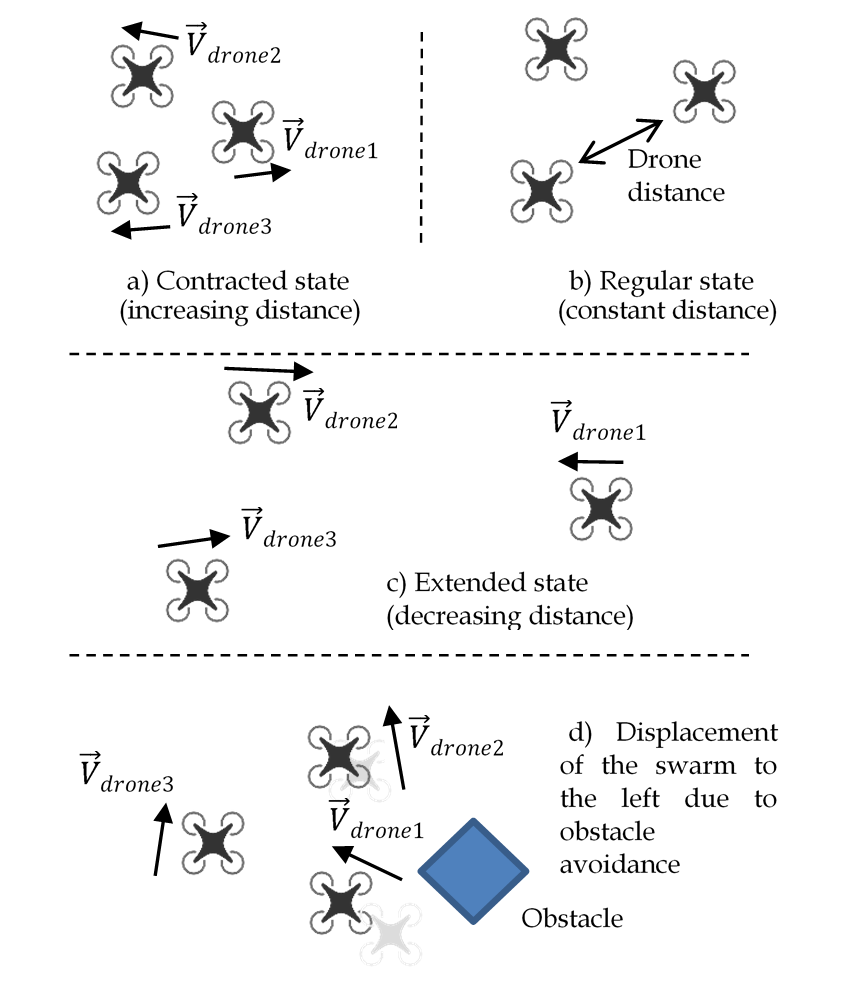}\label{true}
\caption{The information to be presented with the wearable tactile
interface. Contracted (a), regular (b), extended (c) state of the formation, and displacement of the formation (d).}
\label{state}
\end{figure}

During swarm manipulation by the operator, the formation can change its shape, becoming contracted or extended (Fig. \ref{state}(a, b, c). Therefore, the operator should take this information into account, since it contributes to better swarm operation in a cluttered environment. For instance, if the swarm gets too contracted, there is a risk of a collision between the drones. On the other hand, while guiding the formation through the obstacles, the extended state of the swarm can also lead to the collision or to a separation of the swarm to two groups. However, in many cases, the formation state is changing dynamically. In such a scenario, additional real-time information on state propagation direction could be provided to the human operator, in particular, whether the drones are flying away from each other (distance between agents is increasing) or the drones are flying toward each other (distance is decreasing).

\begin{figure}[t]
\centering
\includegraphics[width=0.49\textwidth]{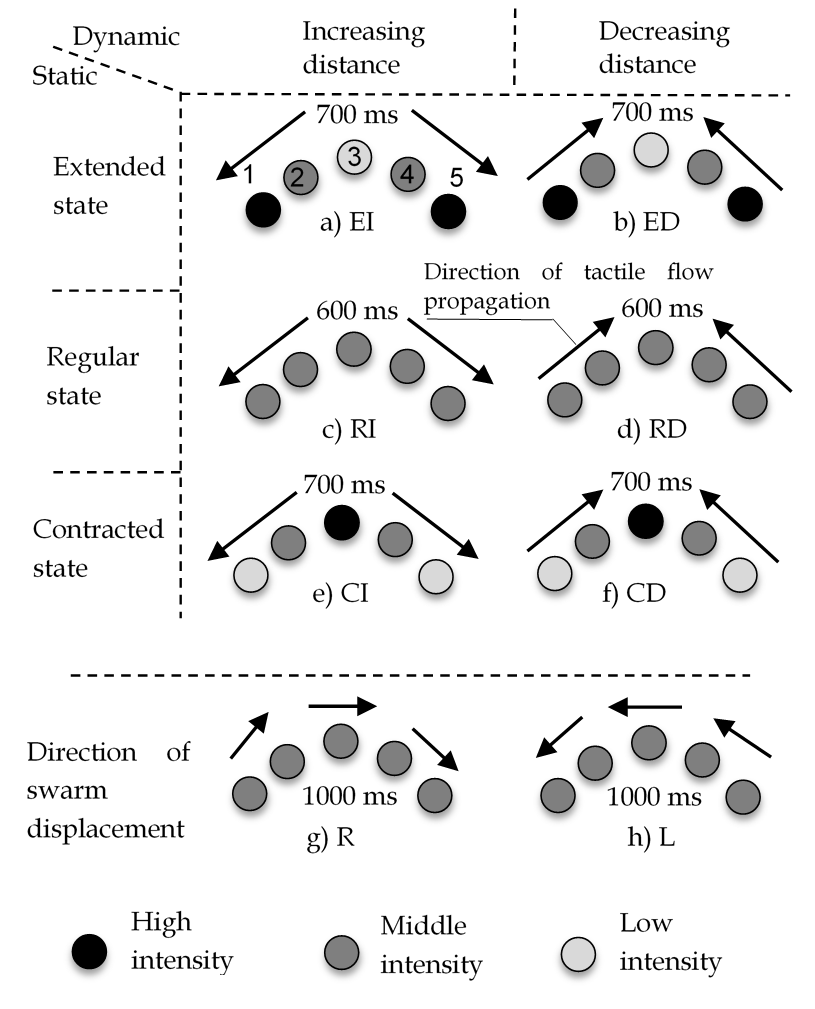}\label{true}
\caption{Tactile patterns for representing the state of the formation in terms of drone-to-drone distance and swarm displacement. Each circle represents finger of a right hand (view from the dorsal side of the hand). The gray scale color represents the intensity of tactor vibration.}
\label{patterns}
\end{figure}

The dynamic change of distance is presented by the tactile flow propagation, e.g., if the distance is increasing, the flow goes from the middle finger to the outer fingers (Fig. \ref{patterns}(a, c, e)), otherwise, the flow goes from the outer fingers to the middle one (Fig. \ref{patterns}(b, d, f)).

The distance between drones is presented by the gradient of the tactor vibration intensity. If the formation is extended, then side vibration motors have a higher intensity than the middle one, see Fig. \ref{patterns}(a, b).

The other swarm state that we propose to present to the operator is the displacement of the center of mass of the swarm to the right or to the left in respect to the direction of motion (Fig. \ref{state}(d)). Due to external factors as obstacles or wind, the swarm could move from the desired direction of motion. The swarm could separate into two groups while avoiding obstacles, which would also lead to the displacement of the center of mass. The direction of displacement is presented with the direction of tactile flow propagation, e.g., when the center of mass is moved to the right with respect to the overall direction of motion, the tactile flow moves from the left finger to the right as shown in Fig. \ref{patterns}(g).

\section{Experiment for Recognition of Tactile Patterns}

\subsection{Participants}
Twenty-two right-handed volunteers (18 males and 4 females, aged 17-36) participated in the experiment. They were given a period for training (5-10 minutes) so that they could get used to the sensations and learn to recognize the signals. All participants positively responded to the device convenience and level of perception.

\subsection{Experiment Condition}
Optimal sensitivity of the skin is achieved at frequencies between 150 and 300 Hz \cite{Jones_2008}. Therefore, for 3 vibration levels, we assigned average frequency values: 150 Hz, 200 Hz, 250 Hz (refer to three grayscale colors shown in Fig. 12). Tactile pulses lasted for 200 or 300 ms depending on the pattern, since distinguishing tactile patterns is easier with stimulus duration in the range of 80 to 320 ms \cite{Jones_2008}.

\subsection{Detection of Multi-modal Patterns}
The experiment was devoted to the detection of multi-modal patterns. The change of distance between drones was modulated by the vector of propagation of tactile stimuli (e.g., if the swarm is extending, firstly the third finger is activated, then, after shut down of the motor on the third finger, the second and fourth fingers are activated, and finally only the first and the fifth ones are vibrating, see Fig. 12(a) for reference). The state of the formation was mapped by the gradient of the vibration intensity (e.g., if the swarm is extended, side fingers have a higher intensity, see Fig. 10(a,b) for reference). To emphasize the direction of the gradient, we introduced different duration of the tactile stimulus. The duration of the tactile pulse in the case of low (150 Hz) and middle (200 Hz) intensity was 200 ms, meanwhile, the duration of the tactile pulses with high (250 Hz) intensity was 300 ms. There was no time interval between the tactile pulses within the same pattern. The total duration of tactile patterns is presented in Fig. 12 and ranged from 600 ms up to 1000 ms.

During the experiment, each pattern was repeated once, and the subject was asked to enter the number of experienced stimuli. Each of the subjects experienced 64 stimuli (8 patterns were repeated 8 times in random order). Time of user response was also recorded.
The results of the user study for the experiment are listed in Table 1. The name of the patterns goes as follows: Extended state, Increasing distance (EI) Fig. 12(a); Extended state, Decreasing distance (ED) Fig. 12(b); Regular state, Increasing distance (RI) Fig. 12(c); Regular state, Decreasing distance (RD) Fig. 12(d); Contracted state, Increasing distance (CI) Fig. 12(e); Contracted state, Decreasing distance (CD) Fig. 12(f); Right displacement (R) Fig. 12(g); Left displacement (L) Fig. 12(h). The diagonal term of the confusion matrix indicates the percentage of the correct responses of participants.

The results of the experiment revealed that all designed tactile patterns were detected by users with an average recognition rate of 76.8\%. Table 1 shows that the distinctive patterns EI, ED, RI, R, and L have higher percentages of recognition and therefore are recommended for the usage in the flight experiment. On the other hand, patterns RD, CI, and CD have lower recognition rates.
One common feature of CI and CD patterns is that they have low vibration intensity of the side fingers.
Therefore, the intensity of the vibration of the fingers number 1 and number 5 (Fig. \ref{glove}) (side fingers) plays a key role in the higher recognition rate.
It can be seen that participants mostly confused patterns CD with RD and patterns CI with RI, while other patterns are distinguished in majority cases. Therefore, it is required to design more distinctive tactile stimuli to improve the recognition rate in some cases. It is important to notice that the direction of tactile flow propagating was distinguished in most cases, both in cases middle-side/side-middle (EI, ED, RI, RD, CI, CD) or left-right/right-left (R, L) direction. Patterns R and L demonstrated the best recognition rates. One reason is that the direction of the tactile flow propagation is easy to recognize. Another potential reason is that patterns R and L have the longest duration. Finally, patterns R and L have a completely different structure - propagation from side to side, apart from all other patterns. Therefore, having six patterns, that have a similar structure (propagation in the middle-side/side-middle direction), could lead to a reduction of the recognition rate.

In order to evaluate the statistical significance of the differences between patterns, we analyzed the results of the user study using single factor repeated-measures ANOVA, with a chosen significance level of $p<0.05$. According to the ANOVA results, there is a statistically significant difference in the recognition rates for the different patterns, $F(7,168)=22.2$, $p=4.3 \cdot 10^{-21}<0.05$. The ANOVA showed that the type of patterns significantly influences the percentage of correct responses.

The paired t-tests showed statistically significant differences between most patterns. For example, there are significant differences between patterns EI and RI ($p=0.023625<0.05$), EI and RD ($p=0.000643<0.05$), EI and CI ($p=7.53\cdot10^{-5}<0.05$), EI and CD ($1.05\cdot10^{-6}<0.05$), EI and R ($p=0.029266<0.05$), EI and L ($p=0.003584<0.05$), ED and RI ($p=0.007042<0.05$) and others. However, the results of paired t-tests between patterns EI and ED, ED and R, ED and L, RI and RD, R and L did not reveal any significant differences, so these patterns have nearly the same recognition rate.

The average time of response, which is the time between the end of pattern execution and the moment when the key is pressed on the keyboard, is slightly different for each participant. From Table 2, we can conclude that participants in the experiment have spent less time to guess pattern R and pattern L. Based on average recognition time, we could conclude that patterns R and L could contribute to more fast and intuitive immersion into the control process, which makes them good candidates for the verification during the flight experiment.
The longest time was 8.92 seconds for Pattern EI. On the other hand, 1.72 seconds is the shortest time period for Pattern R. On average, 3.53 seconds have been spent to response for pattern recognition.

\begin{table}[]
\centering
\caption{Confusion Matrix}
\label{confusion_matrix}
\begin{tabular}{|l|l|l|l|l|l|l|l|l|}
\hline
 & EI & ED & RI & RD & CI & CD & R & L \\ \hline
EI & \textbf{89.8} & 1.1 & 4.0 & 0.0 & 1.7 & 1.1 & 0.0 & 2.3 \\ \hline
ED & 1.1 & \textbf{93.2} & 0.0 & 1.7 & 0.0 & 2.3 & 0.6 & 1.1 \\ \hline
RI & 14.8 & 0.0 & \textbf{76.1} & 1.1 & 5.1 & 1.1 & 0.6 & 1.1 \\ \hline
RD & 0.0 & 21.6 & 1.1 & \textbf{68.8} & 1.1 & 4.5 & 0.6 & 2.3 \\ \hline
CI & 3.4 & 0.6 & 38.1 & 1.7 & \textbf{53.4} & 1.7 & 0.0 & 0.0 \\ \hline
CD & 1.7 & 1.7 & 2.8 & 48.3 & 0.6 & \textbf{40.3} & 1.1 & 0.0 \\ \hline
R & 0.6 & 0.0 & 2.8 & 0.6 & 0.0 & 0.0 & \textbf{95.5} & 0.6 \\ \hline
L & 0.0 & 0.6 & 0.6 & 0.6 & 0.0 & 0.0 & 0.6 & \textbf{97.7} \\ \hline
\end{tabular}
\end{table}

\begin{table}[]
\centering
\caption{Average Time of Recognition Response}
\label{time_of_recognition}
\begin{tabular}{|l|l|l|l|l|l|l|l|l|}
\hline
 & EI & ED & RI & RD & CI & CD & R & L \\ \hline
Time, s & 3.55 & 3.47 & 3.56 & 4.58 & 3.81 & 4.58 & 3.12 & 2.92 \\ \hline
\end{tabular}
\end{table}

\section{Flight Experiment with the SwarmGlove}
To estimate the performance of the SwarmGlove, we set up the flight experiment, in which the user has to navigate the fleet of three Crazyflie 2.0 drones through the obstacles.

\subsection{Role of Tactile Feedback} \label{combination_of_visual_and_tactile_feedback}
As discussed above, the proposed tactile interface could be helpful when the visual feedback of the fleet operator has a poor quality or overloaded with information. On the one hand, communication problems or limited field of view of onboard sensors could lead to the degradation of the visual channel. Additionally, the limited cognitive abilities of the human operator, prevent the user from fully understanding the state of the fleet, especially when the number of drones is high. In such cases, the tactile interface could supplement or even replace the visual feedback.

Considering small size drones, such as Crazyflie 2.0 which can move fast, and a limited flight space (5 m x 5 m x 5 m), the state of the fleet could be changed in a fraction of a second during the experiment. To operate the formation in such an environment, the visual feedback is sufficient, because it is fast and can cover all flight space. Supplementing the visual feedback with the tactile feedback is inefficient in our experimental conditions since it takes up to one second to execute a tactile pattern. We conducted several preliminary flights, providing both visual and tactile feedback to the subjects, but the users relied only on the visual channel.
The other option is to completely replace the visual channel with a tactile sensation. It could be useful when the fleet flies through the areas where it is impossible to acquire or transfer high quality visual information. For this reason, we conducted a flight experiment with only visual feedback and with only tactile feedback. Our hypothesis is based on the assumption that the developed tactile interface could help to navigate the fleet through the blind zones with no visual feedback.

Current experimental conditions do not allow to supplements the visual channel with tactile, due to the reasons discussed above. However, in real life applications, the size of the operational area could be big enough to prevent visual observation of the whole space, and the size of the robots and the size of the formation could lead to a relatively slow change in the fleet state. In such a case, developed tactile feedback could effectively contribute to the visual feedback, by not only replacing but also by supplementing it.

\subsection{Information to be Presented to the Operator}
The next decision that was made was about the parameters of the fleet that have to be reported to the human operator through the tactile interface. As discussed before, for the flight experiment, we use small quadrotors and limited flight space. In such an operational condition, change of the formation shape (increasing or decreasing drone-to-drone distance) could happen very fast, and therefore, it is inefficient to provide slow tactile feedback about it.

\begin{figure}[t]
\centering
\includegraphics[width=0.49\textwidth]{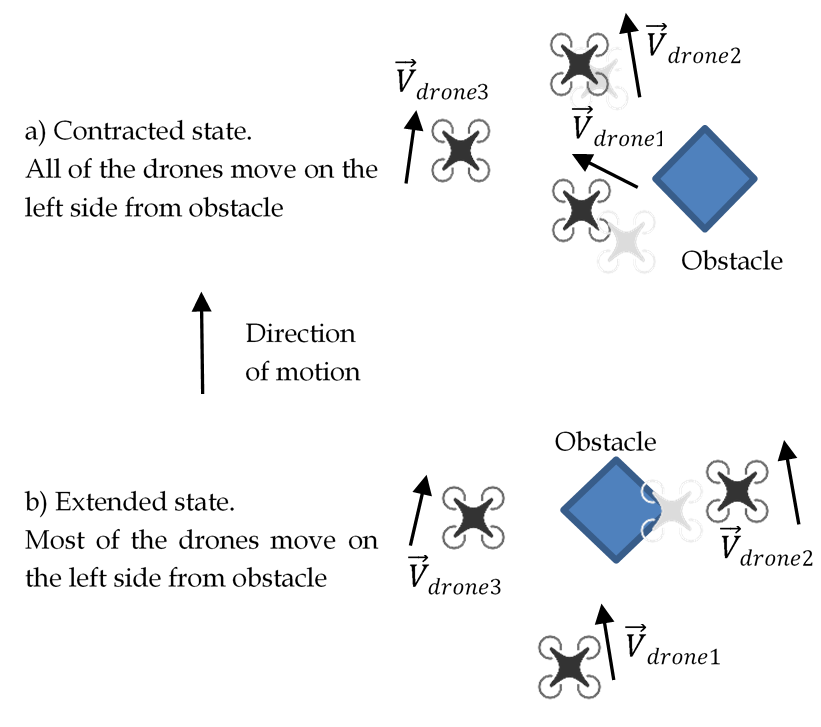}\label{true}
\caption{Information to be presented to the human operator.
Drones avoid obstacles.}
\label{state_fligth_experiment}
\end{figure}

On the other hand, contracted (Fig. \ref{state_fligth_experiment}(a)) or extended (Fig. \ref{state_fligth_experiment}(b)) state of fleet could last for seconds, which makes them applicable candidates for the flight verification. For the experiment, we assume that the formation has a default configuration of the equilateral triangle. We decided that if the area of triangle or distance between any pair of drones is more than 10\% bigger or 10\% less than the default value, then the formation is considered to be in the extended or contracted state, respectively.

Along with the contracted or extended state, it is reasonable to provide the direction of fleet center of mass (CoM) displacement. CoM displacement could happen in both extended or contracted state.
For example, in the contracted state shown in Fig. \ref{state_fligth_experiment}(a), CoM moves to the left with respect to the direction of motion, as far as drone1 and drone2 move to the left from their default positions. Considering the extended state, as shown in Fig. \ref{state_fligth_experiment}(b), CoM moves to the right as far as the drone2 avoiding the obstacle over the right side.

The displacement direction in the contracted state (Fig. \ref{state_fligth_experiment}(a)) is straightforward from the operator point of view, as all drones move collinear with CoM displacement (in Fig. \ref{state_fligth_experiment}(a) drone1 and drone2 move to the left and the CM displaces to the left as well). On the other hand, the displacement direction in the extended state is more complicated to understand, see Fig. \ref{state_fligth_experiment}(b), since the CoM moves on the right, but the majority of the drones go around the left side of the obstacle (the overall goal is to keep the default shape to be able to complete a successful flight mission, formation division is not allowed). To address this complication, we designed tactile feedback patterns to be more intuitive for the operator. To avoid misunderstanding from the user, for the experiment we designed the patterns to inform the user about the recommended direction of hand motion to minimize CoM displacement, rather than the displacement of the CoM itself.

\subsection{Simplified Patterns for the Flight Experiment}
For the next step, we selected which patterns to use to represent contraction, extension, and displacement. 
Initially, we designed the system to be applied for outdoor operation in unstructured environments such as cities, where the fleet moves slowly and the distances are much bigger than indoors. Considering the small flight facilities that were available for this study (5 m x 5 m x 5 m size room), the state of the fleet of three Crazyflies could change rapidly. Therefore, we decided to upgrade the high-quality patterns (EI, ED, R, and L from Section 4.3) and design faster and simpler versions of them for indoor flight test.

\begin{figure}[t]
\centering
\includegraphics[width=0.49\textwidth]{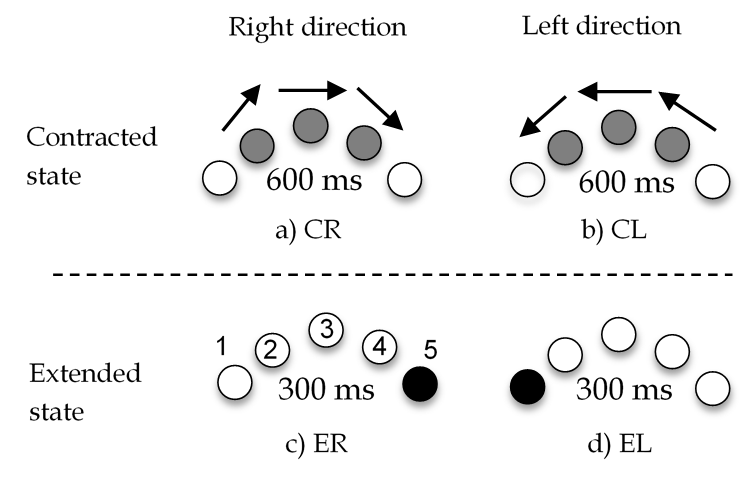}\label{true}
\caption{Simplified tactile patterns for representing the state of the formation in terms of drone-to-drone distance and fleet displacement. Designed for the flight experiment.}
\label{fig:simple_patterns}
\end{figure}

Our goal was to design patterns which will present two types of information: extension/contraction and the direction of motion to prevent the center of mass displacement. Developed multi-modal patterns are presented in Fig. \ref{fig:simple_patterns} (CR – Contracted state, Right Direction; CL – Contracted state, Left Direction; ER – Extended state, Right Direction; EL – Extended state, Left Direction;). For the contracted state, we use three middle fingers (2, 3, 4) and for the extended state, we use side fingers (1 and 5). For the contraction, the direction of the displacement is shown with the tactile flow propagation. For the extension, the direction of the displacement is shown with the right or left finger. Presented patterns are easier to recognize and several times faster than the patterns shown in Fig. \ref{patterns}.
The recognition rate is 100\% (based on 160 trials among 8 participants).

In the case of CR or CL pattern, the best decision is to move the fleet towards the direction of the pattern. For the ER or EL patterns, the best strategy is to move a little bit back (to prevent separation of the fleet) and then move towards the vibrating finger. All of these strategies were presented to the subjects during the training of the user study experiment. As discussed at the end of Section 5.2, the center of mass displacement correlates differently with the proper direction of safety movement. That is why for the patterns CR or CL, the direction of displacement is collinear with the displacement of the center of mass, while for the ER or EL patterns, it is the opposite.

\subsection{Experimental Methodology}
Six right-handed male users (22 to 28 years old) took part in the flight experiment where they were asked to navigate the formation of three Crazyflie 2.0 drones through a labyrinth with obstacles (Fig. \ref{swarmtouch}) using either only visual or only tactile feedback. The state of the formation could be changed due to obstacle avoidance or impedance interlinks as described in the upper sections. The motion is constrained to be two-dimensional on the same height. In order to remove the sound of the drone motors, subjects wore noise-canceling headphones playing white noise. Each participant wore safety glasses. The protocol of the experiment was approved by a Skolkovo Institute of Science and Technology review board, and all participants gave informed consent.

The obstacles used were vertical columns with unlimited height. Participants were not aware of the configuration of the labyrinth beforehand.
The main goal for the participants was to avoid the non-default states of the formation, such as contraction or extension. 
To complicate the visual trial, the obstacles were placed below the flying altitude of the drones and were virtually extended to an unlimited height, thus making it more difficult to visually approximate the distance between a drone and an obstacle.

Training included learning patterns presented in Fig. \ref{fig:simple_patterns}. All possible decision strategies regarding different tactile patterns were presented to the users during the training period.
Training also included guiding the formation through the maze with only tactile feedback first, and with only visual feedback after. 
Regarding the tactile trial, in the default state, no tactile patterns were provided, which meant that it was possible to move forward.

After training, for the experiment, users overcame two different unknown configurations of obstacles, first with tactile and then with visual feedback (two trials with tactile and two with visual feedback in total).


\subsection{Flight Experiment Results}
The initial hypothesis was confirmed. It is possible to navigate the fleet of drones in a cluttered environment using only tactile feedback about the state of the fleet. Users successfully completed the labyrinth in 12 trials, and only two collisions between drones occurred (collision cases were not considered in the statistical analysis presented in Table \ref{table:performance_parameters}; in the case of collision the experiment was repeated). 

As discussed in Section \ref{combination_of_visual_and_tactile_feedback}, for the current experimental conditions, performance with visual feedback is better than with tactile feedback. We compare some of the parameters to understand the behavior of participants better in both cases. The mean values of parameters over all participants are presented in Table \ref{table:performance_parameters}.
The mean path length of the formation centroid is almost two times longer for the tactile feedback, which indicates that with tactile feedback subjects explore the space more actively. The mean velocity with visual feedback is three times faster. Considering the acceleration and jerk of the centroid, it could be included that with tactile feedback, the fleet is guided more smoothly. One of the main metrics is the area of the triangle (formed with actual drone positions) while going through the labyrinth. It is interesting that tactile and visual performance do not differ much in this metric. Therefore, the developed tactile interface does not only allow the possibility to navigate the fleet through cluttered environments but also to do so in a precise manner, maintaining the desired geometry of the formation.

\begin{table}[]
\centering
\caption{Parameters of the Fleet Performance}
\label{table:performance_parameters}
\begin{tabular}{|l|l|l|}
\hline
 & \multicolumn{2}{l|}{\textbf{Feedback type}} \\ \hline
\textbf{Parameters} & Tactile & Visual \\ \hline
Mean centroid length of a path, m & 6.00 & 3.76 \\ \hline
Mean centroid velocity, m/s & 0.08 & 0.23 \\ \hline
Mean centroid acceleration, m/s2 & 0.16 & 0.31 \\ \hline
Mean centroid jerk, m/s3 & 1.03 & 1.92 \\ \hline
Mean of area error, m2 & 0.01 & 0.007 \\ \hline
Mean std. deviation of area error, m2 & 0.009 & 0.006 \\ \hline
Mean maximum of area error, m2 & 0.039 & 0.028 \\ \hline
\end{tabular}
\end{table}

\begin{figure}[t]
\centering
\includegraphics[width=0.49\textwidth]{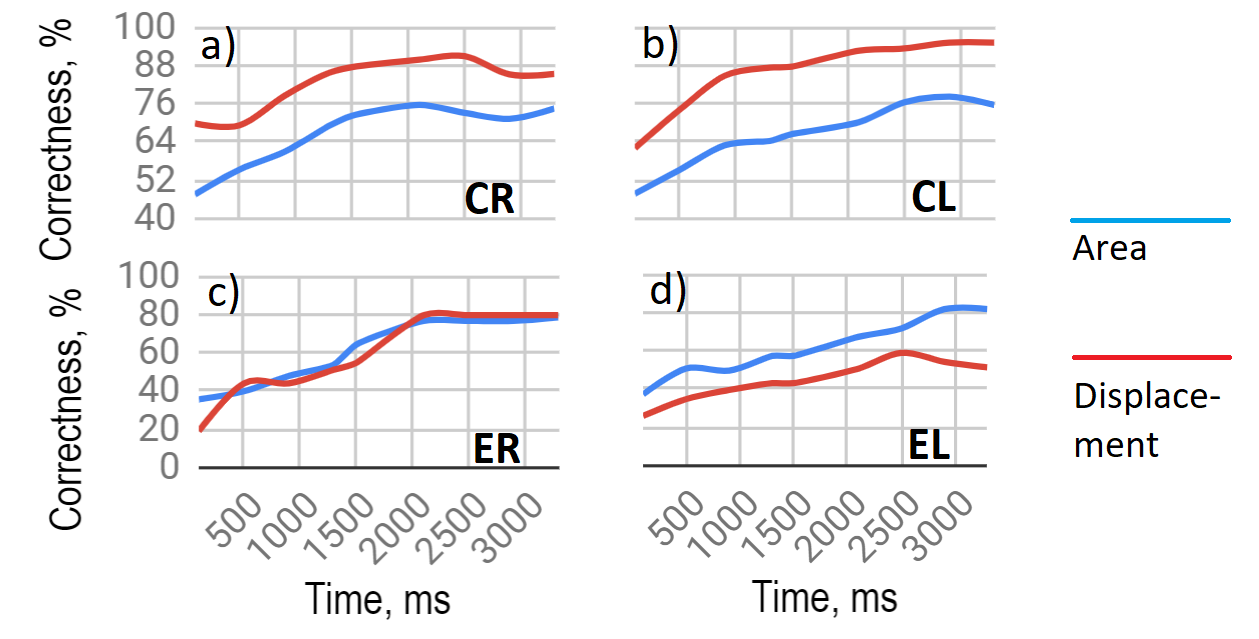}\label{true}
\caption{Percentage of proper correction (returning area to the default value and decrease of displacement) of fleet state after the start of pattern execution.}
\label{fig:reaction_correctness}
\end{figure}

Considering more closely the behavior of the users with respect to the executed patterns, we investigated the fleet behavior right after the patterns were executed (see Fig. \ref{fig:reaction_correctness}).
In the example of state change shown in Fig. \ref{state_fligth_experiment}(a) for example (contraction and displacement to the left), the user receives a CL pattern (Fig. \ref{fig:simple_patterns}). Then, as the formation is guided to the left the fleet centroid should move to the left and the area should increase back to the default value. 
We compared the area and centroid displacement at the current time (for the time interval 0-3300 ms after the start of each pattern execution) with the corresponding values at the time when the pattern started.
Creating such comparisons for all patterns helped us to evaluate the correctness and duration of human operator response. The evaluation was performed for all participants.
It is possible to see from Fig. \ref{fig:reaction_correctness} that, in general, the correctness of operator decisions reaches 75-80\% after 2-3 seconds after pattern execution. CR/CL patterns work better for the displacement, and the overall performance of CR/CL is higher than ER/EL. It might be because of a more simple behaviour strategy for CR/Cl patterns (discussed in Section 5.3).

\begin{figure}[t]
\centering
{\includegraphics[width=0.49\columnwidth]{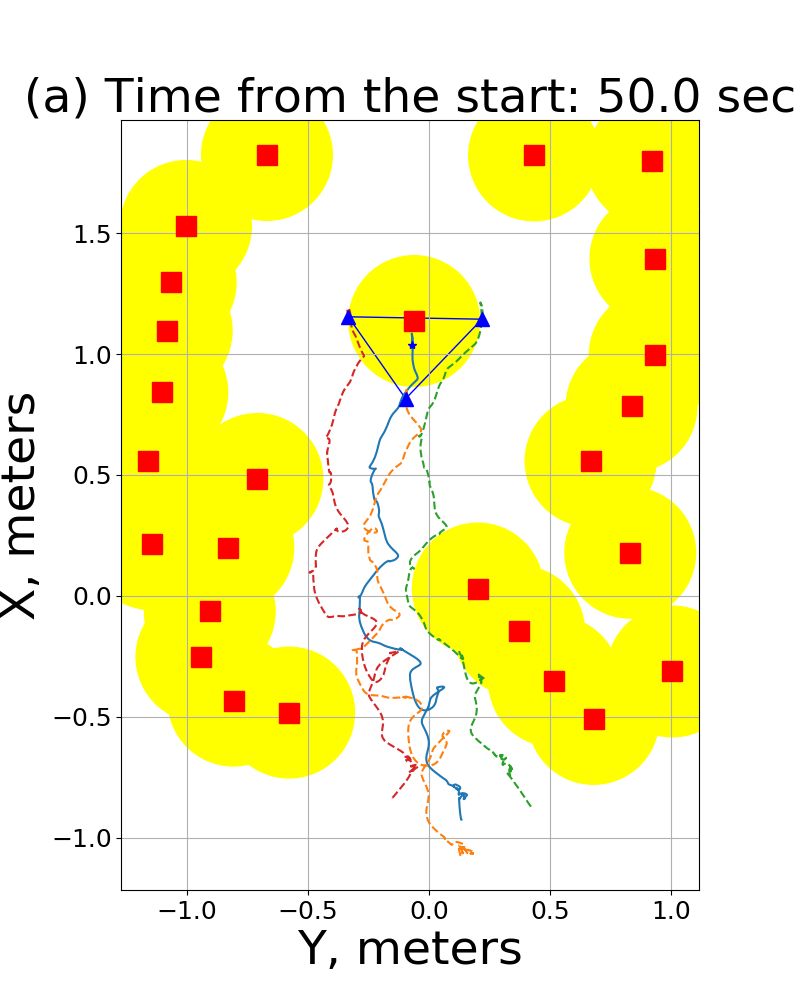}\label{3}}
{\includegraphics[width=0.49\columnwidth]{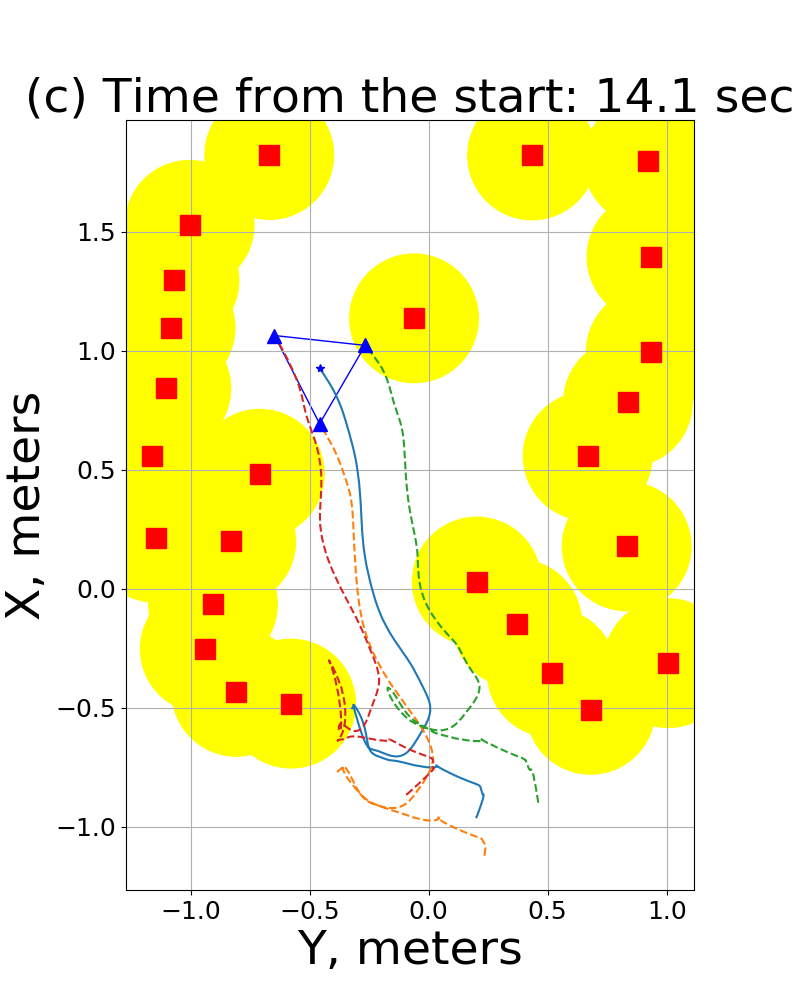}\label{4}}
{\includegraphics[width=0.49\columnwidth]{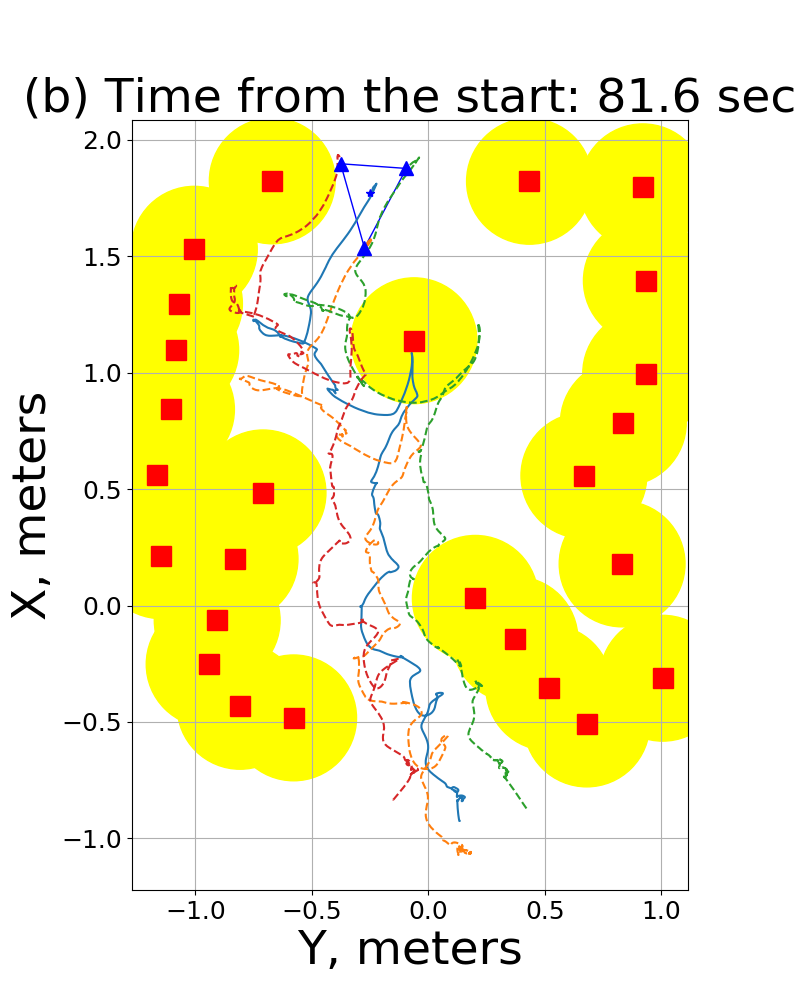}\label{5}}
{\includegraphics[width=0.49\columnwidth]{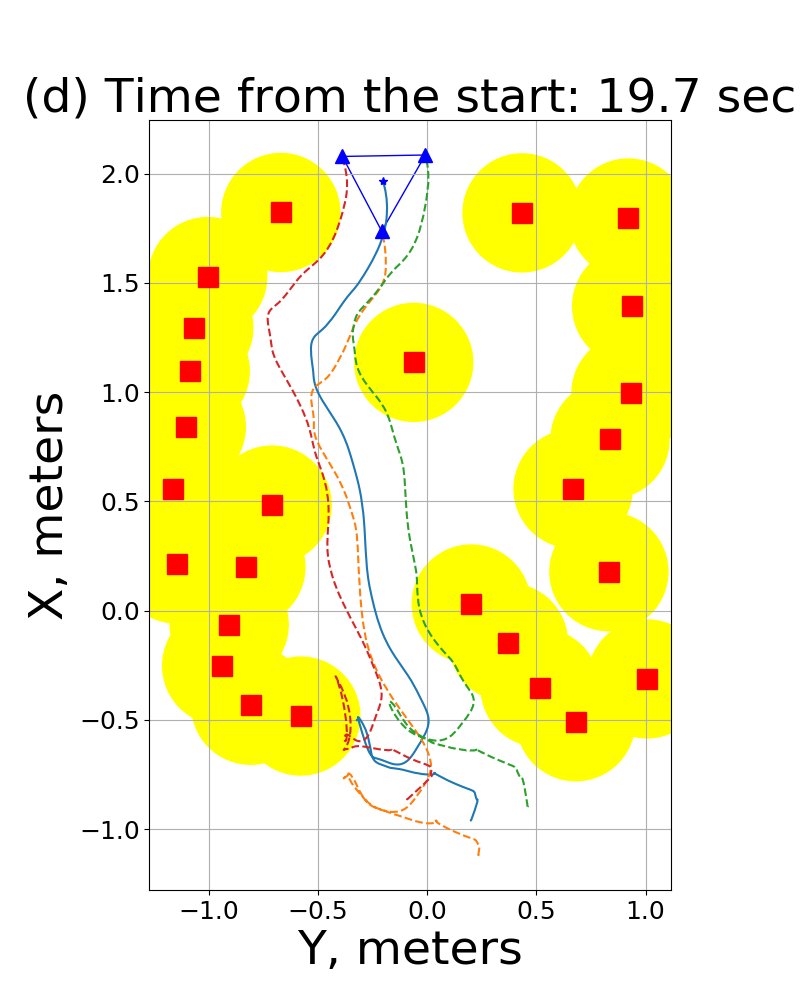}\label{6}}
\caption{A top view of trajectories of drones and centroid while being manipulated thought the maze. Left column of two pictures (a-b) represents formation of three drones navigation with the help of pure \textit{tactile feedback}. Two pictures on the right (c-d) - navigation with pure \textit{visual feedback}. Solid line is the trajectory of the fleet centroid. Dashed lines are actual drones trajectories. Red squares represent real obstacles with yellow safety zones. Formation shape is depicted with blue triangle. }
\label{fig:labyrinth}
\end{figure}

Example of trajectories during the experiment are shown in Fig. \ref{fig:labyrinth}. Drones trajectories (dashed lines) are represented in $XY$-plane (from the top view). The fleet flies among obstacles (small red squares). Each of the obstacles is surrounded by the yellow cylindrical safety zone. The union of all these cylindrical obstacles vicinity defines the area, where drones cannot fly. The solid blue line represents the fleet central point path.

It could be noticed in the left column of Fig. \ref{fig:labyrinth} (tactile trial) that centroid trajectory has several turns near the obstacle vicinities. At these moments a human operator receives tactile patterns that help him to understand that the fleet is located near the obstacles and also provides information on how to control drones in order to avoid collisions and reach the finish point.

\section{Conclusion and Future Works}
We have proposed a novel system SwarmTouch which integrates impedance control and a tactile glove for intuitive swarm control by an operator. The impedance links between agents and an intelligent obstacle avoidance algorithm allow the swarm to not only generate a safe trajectory but also to perform a smooth motion. We also designed the tactile patterns for the glove and conducted experiments to determine more recognizable patterns. The flight experiments demonstrated accurate guidance of the swarm using tactile sensation, which is close to visual feedback navigation in terms of geometry maintenance. The experimental data are available at \underline{\url{http://doi.org/10.5281/zenodo.3256614}}.

To track positions of quadrotors with Global Navigation Satellite System (GNSS) is accurate enough in most cases. However, it could be hard to track small hand motions with GPS. Therefore, the current work could be extended towards the development of the local positioning system for hand tracking. The other option is to replace the hand position and velocity as a control input with something else. The alternative method could be to use an inclination of hand (standard inertial-measurement unit (IMU) could measure all necessary information) instead of the hand position for the control input.  It is also possible to use a joystick, rather than a glove, as a control input device.

Due to its mobility and spatial distribution, the fleet of quadrotors could be the first responder for different kinds of emergencies, such as fire, earthquake or flood. To gather information about a suffering area is a crucial task for first responders. Monitoring of the progress of disaster recovery is also important because an emergency can have a dynamically changing environment. Navigation of a swarm in a city environment, with multi-story buildings or even skyscrapers could be a challenging task.
Maintaining the default geometry of the formation is a reasonable requirement for real-life applications when data must be gathered evenly or communication within the formation is necessary. 
As a result, SwarmTouch could contribute to a faster response to high risk and uncontrolled situations and a higher level of awareness of a swarm's surroundings for the operator.

\ifCLASSOPTIONcaptionsoff
  \newpage
\fi



%
\bibliographystyle{IEEEtran}
\bibliography{Sk}

\begin{IEEEbiography}
[{\includegraphics[width=1in,height=1.25in,clip,keepaspectratio]{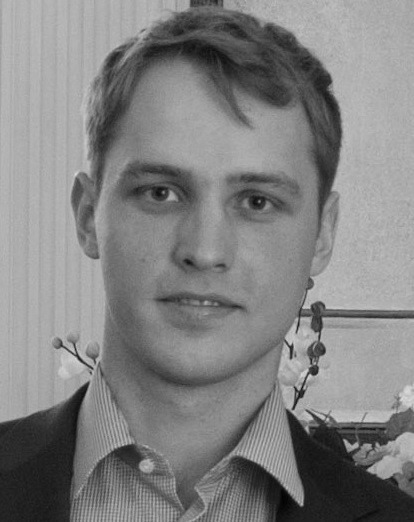}}]
{Evgeny Tsykunov}
received the MS degree in computer science from Skolkovo Institute of Science and Technology, Moscow, Russian Federation in 2016. He is currently working towards the Ph.D. degree in the Intelligent Space Robotics Laboratory, Skolkovo Institute of Science and Technology. 
His research interests include developing control algorithms and interfaces for the human-swarm interaction.
\end{IEEEbiography}

\vskip -2\baselineskip plus -1fil

\begin{IEEEbiography}
[{\includegraphics[width=1in,height=1.25in,clip,keepaspectratio]{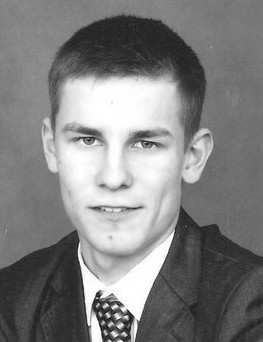}}]
{Ruslan Agishev}
received a Bachelor degree in computer engineering from Moscow Institute of Physics and Technology, Russia, in 2017. He received the MS degree in the Department of Space and Engineering Systems, Skolkovo Institute of Science and Technology.
His research interests include robotics and mathematics, focusing on UAVs control and navigation, path planning algorithms.
\end{IEEEbiography}

\vskip -2\baselineskip plus -1fil

\begin{IEEEbiography}
[{\includegraphics[width=1in,height=1.25in,clip,keepaspectratio]{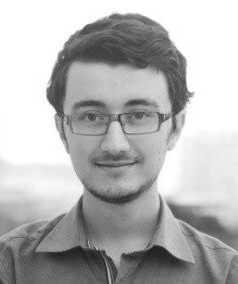}}]
{Roman Ibrahimov}
received the BSc degree in systems engineering from ADA University, Azerbaijan, in 2018. He was an exchange student at METU, Turkey, in 2016 and ITMO University, Russia, in 2017. He is now a first-year MS student at the Department of Space and Engineering Systems, Skoltech, Russia. His research interests include, but are not limited to, robotics and control systems.
\end{IEEEbiography}

\vskip -2\baselineskip plus -1fil

\begin{IEEEbiography}
[{\includegraphics[width=1in,height=1.25in,clip,keepaspectratio]{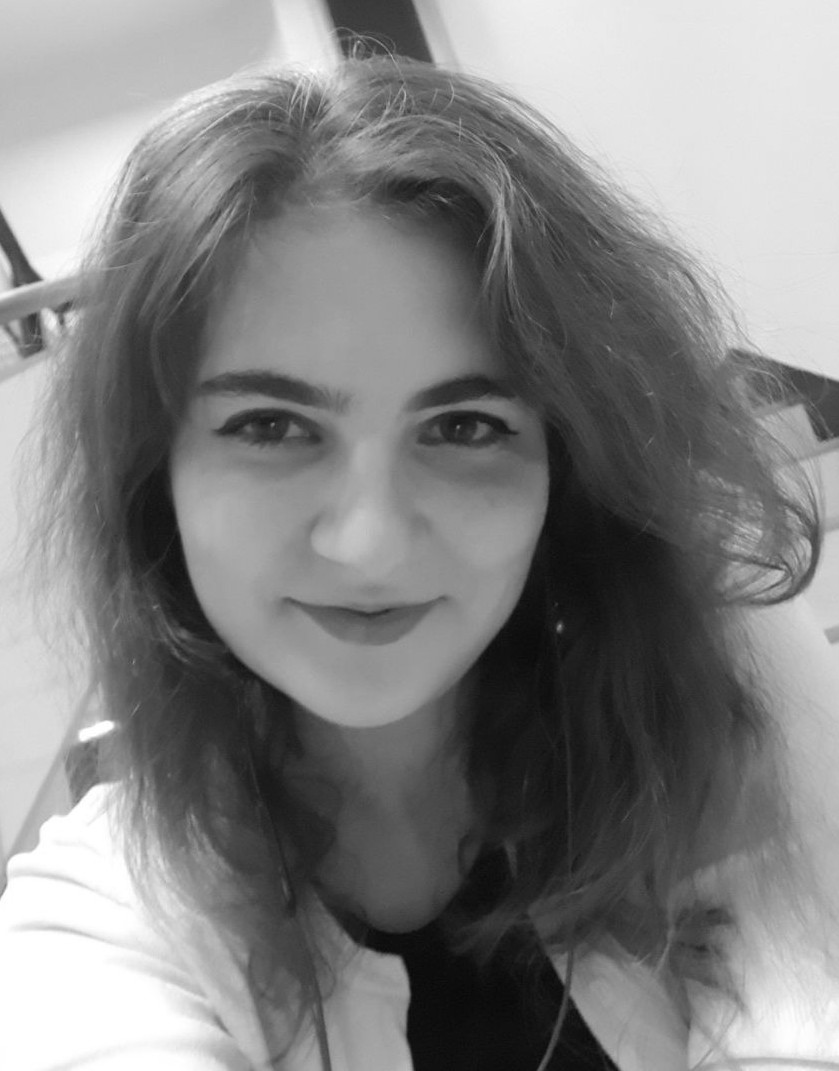}}]
{Labazanova Luiza} 
is currently working toward MS degree in the Department of Space and Engineering Systems, Skolkovo Institute of Science and Technology. In addition, she was a visiting student at the Department of Informatics, University of Electro-Communications, Japan.  Her research interests include robotics and haptics, focusing on wearable devices and mechanoreceptors stimulation.
\end{IEEEbiography}

\vskip -2\baselineskip plus -1fil

\begin{IEEEbiography}
[{\includegraphics[width=1in,height=1.25in,clip,keepaspectratio]{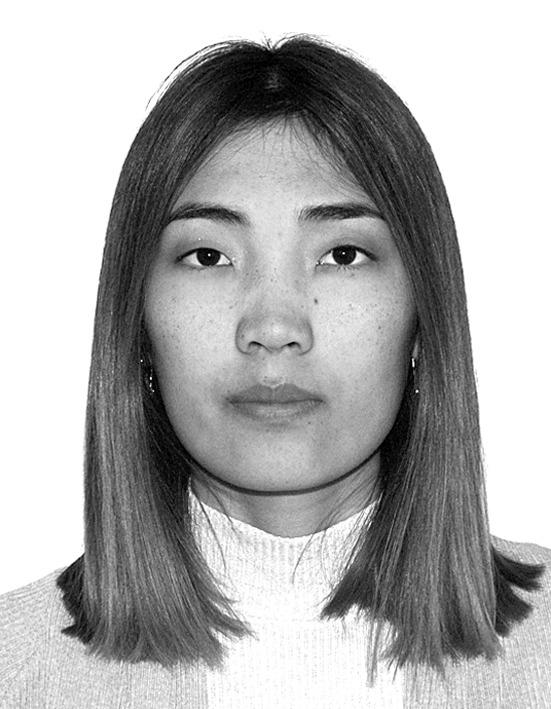}}]
{Akerke Tleugazy}
received the Bachelor Degree in Electrical and Electronics Engineering from the Nazarbayev University in Astana, Kazakhstan, in 2017. Currently, she is the 2nd year Master student in the Space and Engineering Systems, Skolkovo Institute of Science and Technology, Moscow, Russian Federation. Her research interests include haptic devices, robotics, virtual reality.
\end{IEEEbiography}

\vskip -2\baselineskip plus -1fil

\begin{IEEEbiography}
[{\includegraphics[width=1in,height=1.25in,clip,keepaspectratio]{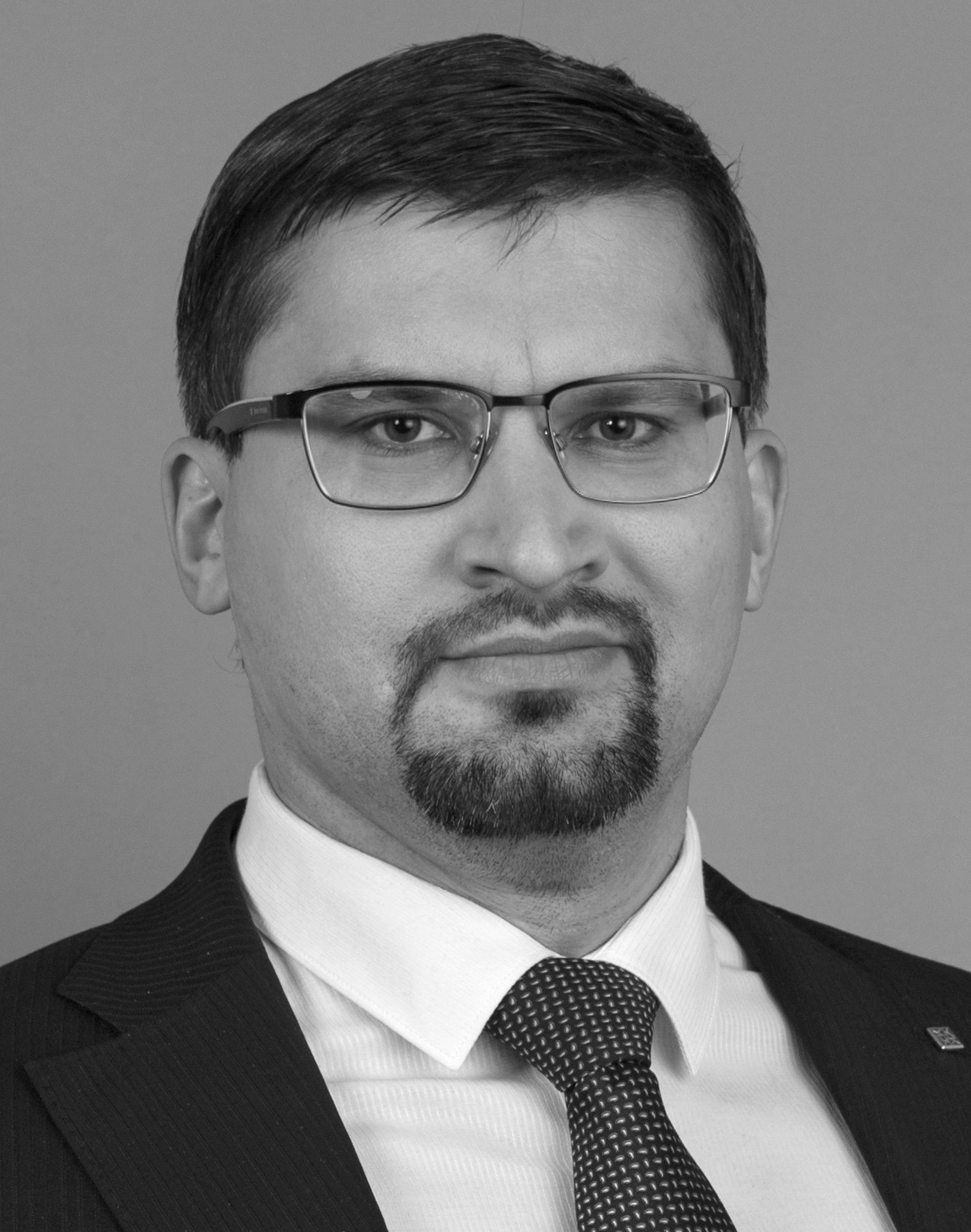}}]
{Dzmitry Tsetserukou}
received the Ph.D. degree in Information Science and Technology from the University of Tokyo, Japan, in 2007. From 2007 to 2009, he was a JSPS Post-Doctoral Fellow at the University of Tokyo. He worked as Assistant Professor at the Electronics-Inspired Interdisciplinary Research Institute, Toyohashi University of Technology from 2010 to 2014. From August 2014 he works at Skolkovo Institute of Science and Technology as Head of Intelligent Space Robotics Laboratory. Dzmitry is a member of the Institute of Electrical and Electronics Engineers (IEEE) since 2006 and the author of over 70 technical publications, 3 patents, and a book.
His research interests include wearable haptic and tactile displays, robotic manipulator design, telexistence, human-robot interaction, affective haptics, virtual and augmented reality. 
Dzmitry is the winner of the Best Demonstration Award (Bronze prize, AsiaHaptics 2018), Laval Virtual Award (ACM Siggraph 2016), Best Presentation Award (IRAGO 2013).
\end{IEEEbiography}




\end{document}